\RequirePackage{fix-cm}
\documentclass[smallextended]{svjour3}       
\smartqed  
\usepackage[table]{xcolor}
\usepackage{xcolor}
\usepackage{graphicx}
\usepackage{amsmath,amssymb}
\usepackage[numbers,authoryear]{natbib}
\bibpunct[,]{(}{)}{;}{a}{}{,}

\def\vector#1{\mbox{\boldmath $#1$}}
\begin{document}

\title{Bayesian sparse convex clustering via global-local shrinkage priors}

\author{Kaito Shimamura         \and
        Shuichi Kawano
}

\institute{K. Shimamura  \at
              NTT Advanced Technology Corporation,
              Muza Kawasaki Central Tower, 1310 Omiya-cho, Saiwai-ku, Kawasaki-shi, Kanagawa 212-0014, Japan.\\
              Graduate School of Informatics and Engineering, The University of Electro-Communications, 1-5-1 Chofugaoka, Chofu-shi, Tokyo 182-8585, Japan.
              \email{kaito.shimamura@ai.lab.uec.ac.jp}
           \and
           S. Kawano \at
             Graduate School of Informatics and Engineering, The University of Electro-Communications, 1-5-1 Chofugaoka, Chofu-shi, Tokyo 182-8585, Japan.
             \email{skawano@ai.lab.uec.ac.jp}
}

\date{Received: date / Accepted: date}

\maketitle

\begin{abstract}
  Sparse convex clustering is to cluster observations and conduct variable selection simultaneously in the framework of convex clustering.
  Although a weighted $L_1$ norm is usually employed for the regularization term in sparse convex clustering, its use increases the dependence on the data and reduces the estimation accuracy if the sample size is not sufficient.
  To tackle these problems, this paper proposes a Bayesian sparse convex clustering method based on the ideas of Bayesian lasso and global-local shrinkage priors.
  We introduce Gibbs sampling algorithms for our method using scale mixtures of normal distributions.
  The effectiveness of the proposed methods is shown in simulation studies and a real data analysis.
\end{abstract}

\section{Introduction}

Cluster analysis is an unsupervised learning method aimed at assigning observations to several clusters so that similar individuals belong to the same group.
It is widely used in such research fields as biology and genomics, as well as many other fields of science.
In general, conventional clustering methods such as $k$-means clustering, hierarchical clustering, and the Gaussian mixture model are instable due to non-convex optimization.

Convex clustering proposed by \cite{Hocking_2011} searches for the centers of all clusters simultaneously with allocating individuals to the clusters.
Convex relaxation ensures that it achieves a unique global optimum regardless of the initial values.
Estimates can be obtained by solving a regularization problem, which is similar to sparse regularization in regression analysis.
However, convex clustering does not work well if the data contain a large amount of irrelevant features.

Sparse regularization is used to exclude irrelevant information.
\citet{wang2018sparse} proposed sparse convex clustering to perform convex clustering and variable selection simultaneously.
Sparse convex clustering estimates sparse models by using the $L_1$ norm in addition to the regularization term of the convex clustering.
Also, \citet{wang2018sparse} used the $L_1$ norm for the convex clustering penalties, where the penalty was assumed to be different weights according to individual and feature.
However, it was pointed out by \citet{griffin2011bayesian} that the penalty used in sparse convex clustering depends on the data, which may lead to model estimation accuracy degradation when the sample size is small.

Our proposed methods overcome the problem that penalties in  sparse convex clustering depend heavily on weight.
In particular, with these methods, even when the sample size is small, estimation is possible without depending on the weight.
To propose a method, we first introduce a Bayesian formulation of sparse convex clustering, and then propose a Bayesian sparse convex clustering based on a global-local (GL) prior distribution.
As the GL prior, we consider three types of distributions: a normal-exponential-gamma distribution \citep{griffin2005alternative}, a horseshoe distribution \citep{carvalho2010horseshoe}, and a Dirichlet--Laplace distribution \citep{bhattacharya2015dirichlet}.
The Gibbs sampling algorithm for our proposed models is derived by using scale mixtures of normal distributions \citep{andrews1974scale}.

The rest of this paper is organized as follows.
Section 2 focuses on the convex clustering method.
In Section 3, we propose a Bayesian formulation of sparse convex clustering.
In Section 4, we propose a Bayesian convex clustering method with GL shrinkage prior distributions.
The performances of the proposed methods are compared with those of the existing method by conducting a Monte Carlo simulation in Section 5 and a real data analysis in Section 6.
Concluding remarks are given in Section 7.

\section{Preliminaries}

In this section, we describe convex clustering.
This is a convex relaxation of such clustering methods as $k$-means clustering and hierarchical clustering.
The convexity overcomes the instability of conventional clustering methods.
In addition, we describe sparse convex clustering which simultaneously clusters observations and performs variable selection.

\subsection{Convex clustering}

Let $X \in \mathbb{R}^{n\times p}$ be a data matrix with $n$ observations and $p$ variables, and $\vector{x}_i\ ( i=1,2,\cdots,n)$ be the $i$-th row of $X$.
Convex clustering for these $n$ observations is formulated as the following minimization problem using an $n \times p$ feature matrix $ A=(\vector{a}_1,\cdots,\vector{a}_n)^T $:
\begin{eqnarray}
  \min_A
  \frac{1}{2}\sum_{i=1}^n
  {\|\vector{x}_i-\vector{a}_i \|}_2^2
  +
  \gamma \sum_{i_1<i_2}
  {\|\vector{a}_{i_1}-\vector{a}_{i_2} \|}_q
  ,
  \label{cvc}
\end{eqnarray}
where $\vector{a}_i$ is a $p$-dimensional vector corresponding to $\vector{x}_i$, $\|\cdot\|_q$ is the $L_q$ norm of a vector, and $\gamma\ (\geq0)$ is a regularization parameter.
If $\hat{\vector{a}}_{i_1}=\hat{\vector{a}}_{i_2}$ for the estimated value $\hat{\vector{a}}_i$, then the $i_1$-th individual and $i_2$-th individual belong to the same cluster.
The $\gamma$ controls the number rows of $\hat{A} = (\hat{\vector{a}}_1, \cdots, \hat{\vector{a}}_n)^T$ that are the same, which determines the estimated number of clusters.
Both $k$-means clustering and hierarchical clustering are equivalent to considering the $L_0$ norm for the second term in the problem ($\ref{cvc}$), which becomes a non-convex optimization problem \citep{Hocking_2011}.
Convex clustering can be viewed as a convex relaxation of $k$-means clustering and hierarchical clustering.
This convex relaxation guarantees that a unique global minimization is achieved.

\begin{figure}
  \begin{center}
    \includegraphics[bb=0 0 960 960 ,width=90mm]{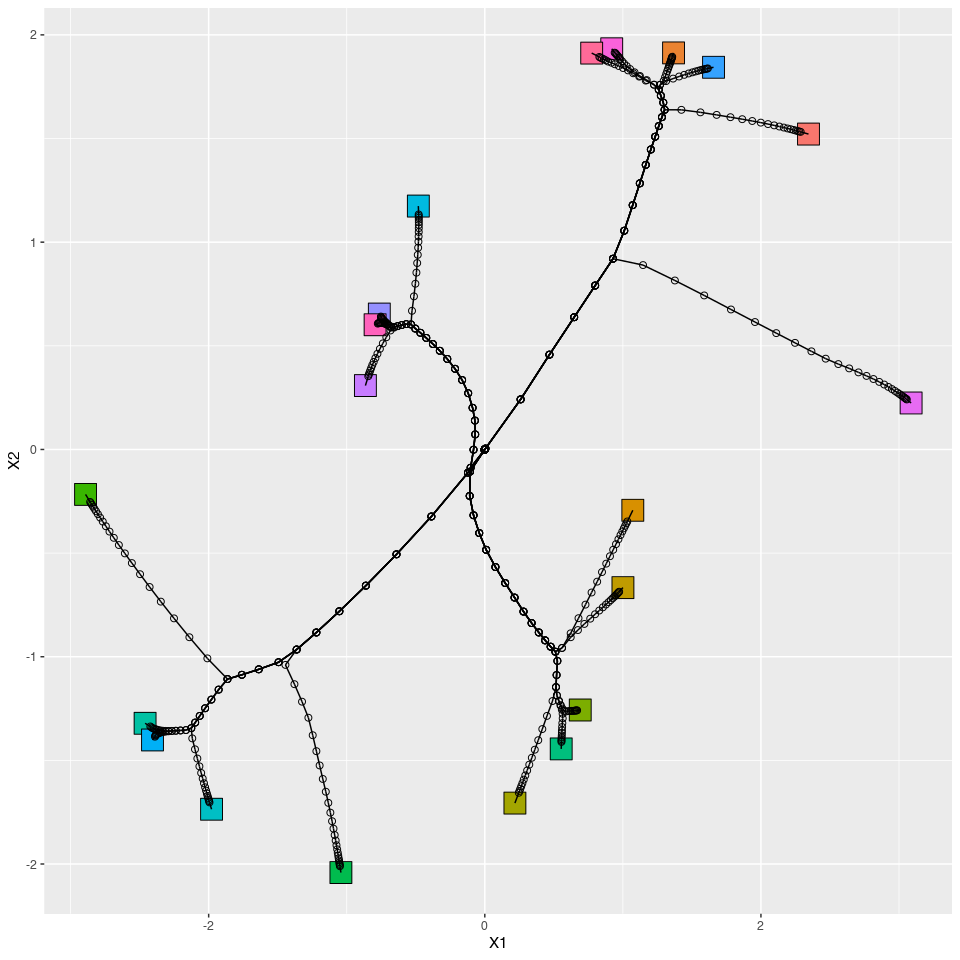}
  \end{center}
  \caption{
  A cluster path for two interlocking half-moons.
  The colored squares are 20 observations and the circles are convex clustering estimates for different regularization parameter values.
  Among the estimates of the same observation, the lines connect the estimates whose values of the regularization parameter are close.
  }
  \label{halfmoons_cvcpath}
\end{figure}

\citet{Hocking_2011} proposed using a cluster path to visualize the steps of clustering.
A cluster path can be regarded as a continuous regularization path \citep{efron2004least} of the optimal solution formed by changing $\gamma$.
Figure $\ref{halfmoons_cvcpath}$ shows the cluster path of two interlocking half-moons described in Section \ref{half_moons}.
A cluster path shows the relationship between values of the regularization parameter and estimates of the feature vectors.
The estimates exist near the corresponding observations when the value of the regularization parameter is small, while the estimates concentrate on one point when the value is large.
The characteristics of the data can be considered from the grouping order and positional relationship of the estimates.

\subsection{Sparse convex clustering}

In conventional convex clustering, when irrelevant information is included in the data, the accuracy of estimating clusters tends to be low.
Sparse convex clustering \citep{wang2018sparse}, on the other hand, is an effective method for such data, as irrelevant information can be eliminated using sparse estimation.

Sparse convex clustering considers the following optimization problem:
\begin{eqnarray}
\min_A \frac{1}{2}\sum_{i=1}^n
{\|\vector{x}_i-\vector{a}_i \|}_2^2
+
\gamma_1 \sum_{(i_1,i_2)\in\mathcal{E}}
w_{i_1,i_2}
{\|\vector{a}_{i_1}-\vector{a}_{i_2} \|}_q
+
\gamma_2
\sum_{j=1}^p u_j \|\vector{a}_{\cdot j}\|_1
,
\label{scvc}
\end{eqnarray}
where
$\gamma_1\ (\geq 0)$ and $\gamma_2\ (\geq 0)$ are regularization parameters, $w_{i_1,i_2}\ (\geq 0)$ and $u_j\ (\geq 0)$ are weights, $q \in \{1, 2, \infty\}$,
$\mathcal{E}=\{(i_1,i_2);  w_{i_1,i_2} \neq 0, i_1<i_2\}$,
and
$\vector{a}_{\cdot j} = (a_{1j},\cdots,a_{nj})^T$ is a column vector of the feature matrix $A$.
The third term imposes a penalty similar to group lasso \citep{yuan2006model} and has the effect that $\|\hat{\vector{a}}_{\cdot j}\|_1 = 0$.
When $\|\hat{\vector{a}}_{\cdot j}\|_1 = 0$, the $j$-th column of $X$ is removed from the model, which is variable selection.
$\gamma_1$ and $w_{i_1i_2}$ adjust the cluster size, whereas $\gamma_2$ and $u_j$ adjust the number of features.
The weight $w_{i_1,i_2}$ plays an important role in imposing a penalty that is adaptive to the features.
\cite{wang2018sparse} used the following weight parameter:
\begin{equation*}
  w_{i_1, i_2}
  =\iota_{i_1, i_2}^{m}
  \exp \left\{
  -\phi\left\|\vector{x}_{i_1}-\vector{x}_{i_2}\right\|_2^2
  \right\}
  ,
\end{equation*}
where $\iota_{i_1, i_2}^{m}$ equals $1$ if the observation $\vector{x}_{i_1}$ is included among the $m$ nearest neighbors of the observation $\vector{x}_{i_2}$, and is $0$ otherwise. This choice of weights works well for a wide range of $\phi$ when $m$ is small.
In our numerical studies, $m$ is fixed at $5$ and $\phi$ is fixed at $0.5$, as in \cite{wang2018sparse}.

Similar to the adaptive lasso \citep{zou2006adaptive} in a regression problem, the penalty for sparse convex clustering can be adjusted flexibly by using weight parameters.
However, it was shown by \cite{griffin2011bayesian} that such penalties are strongly dependent on the data.
In particular, the accuracy of model estimation may be low due to the data, such as because the number of samples is small.

\section{Bayesian formulation of sparse convex clustering}

By extending sparse convex clustering to a Bayesian formulation, we may use the entire posterior distribution to provide a probabilistic measure of uncertainty.

\subsection{Bayesian sparse convex clustering}

In this section, we reformulate sparse convex clustering as a Bayesian approach.
Similar to Bayesian lasso \citep{Park_2008}, which extends lasso to a Bayesian formulation, we regard regularized maximum likelihood estimates as MAP estimates.

We consider the following model:
\[
\vector{x}=\vector{a}+\vector{\varepsilon}
,
\]
where $\vector{\varepsilon}$ is a $p$-dimensional error vector distributed as $\mbox{N}_p(\vector{0}_p, \sigma^2I_p)$,
$\vector{a}$ is a feature vector, and $\sigma^2\ (>0)$ is a variance parameter.
Then, the likelihood function is given by
\[
  f(X|A,\sigma^2)
  =
  \prod_{i=1}^n
  (2\pi\sigma^2)^{-p/2}
  \exp\left\{
  -\frac{\|\vector{x}_i-\vector{a}_i\|^2_2}
  {2\sigma^2}
  \right\}
  .
\]
Next, we specify the prior distribution of feature matrix $A$ as
\begin{eqnarray}
  \pi(A|\sigma^2)
  &\propto&
  (\sigma^2)^{-(\#\mathcal{E}+p)/2}
  \exp\left\{
  -\frac{\lambda_1}{\sigma}
  \sum_{(i_1,i_2)\in\mathcal{E}}
  w_{i_1,i_2}
  \|\vector{a}_{i_1}-\vector{a}_{i_2}\|_2
  \right\}
  \nonumber \\&&
  \times
  \exp\left\{
  -\frac{\lambda_2}{\sigma}
  \sum_{j=1}^p
  u_j
  \|\vector{a}_{\cdot j}\|_2
  \right\}
  ,
  \label{prior_bscvc}
\end{eqnarray}
where $\lambda_1\ (>0),\ w_{i_1,i_2}\ (>0),\ \lambda_2\ (>0),\ u_j\ (>0)$ are hyperparameters, $\mathcal{E} = \{(i_1,i_2): 1 \leq i_1<i_2\leq n\}$, and $\# \mathcal{E}$ is the number of elements in $\mathcal{E}$.
Note that $\lambda_1$ and $\lambda_2$ correspond to $\gamma_1$ and $\gamma_2$ in (\ref{scvc}).
This prior distribution is an extension of Bayesian group lasso in linear regression models \citep{xu2015bayesian}.
The estimate of a specific sparse convex clustering corresponds to the MAP estimate in the following joint posterior distribution:
\begin{eqnarray}
  \pi(A, \sigma^2 | X)
  &\propto&
  f(X|A,\sigma^2)\pi(A|\sigma^2)\pi(\sigma^2)
  \nonumber
  \\
  &\propto&
  (\sigma^2)^{-(np+\#\mathcal{E}+p)/2}
  \exp\left\{
  -\frac{1}
  {2\sigma^2}
  \|\vector{x}_i-\vector{a}_i\|^2_2
  \right\}
  \nonumber
  \\&&
  \times
  \exp\left\{
  -\frac{\lambda_1}{\sigma}
  \sum_{(i_1,i_2)\in\mathcal{E}}
  w_{i_1,i_2}
  \|\vector{a}_{i_1}-\vector{a}_{i_2}\|_2
  \right\}
  \nonumber
  \\&&
  \times
  \exp\left\{
  -\frac{\lambda_2}{\sigma}
  \sum_{j=1}^p
  u_j
  \|\vector{a}_{\cdot j}\|_2
  \right\}
  \pi(\sigma^2)
  ,
  \label{bscvc_prior_orig}
\end{eqnarray}
where $\pi(\sigma^2)$ is the non-informative scale-invariant prior $\pi(\sigma^2)=1/\sigma^2$ or inverse-gamma prior $\pi(\sigma^2)=\mbox{IG}(\nu_1/2,\eta_0/2)$. An inverse-gamma probability density function is given by
\begin{eqnarray}
  \mbox{IG}(x|\nu,\eta)
  =
  \frac{\eta^\nu}{\Gamma(\nu)
  x^{-(\nu+1)}}
  \exp\left\{
  -\frac{\eta}{x}
  \right\}
  ,
  \label{inverse_gamma}
\end{eqnarray}
where $\nu\ (>0)$ is a shape parameter, $\eta\ (>0)$ is a scale parameter, and $\Gamma(\cdot)$ is the gamma function.

We obtain estimates of each parameter by applying the MCMC algorithm with Gibbs sampling.
Therefore, it is necessary to derive the full conditional distribution for each parameter.
Because it is difficult to derive full conditional distributions from (\ref{bscvc_prior_orig}),
we derive a hierarchical representation of the prior distribution.
First, the prior distribution $\pi(A|\sigma^2)$ is rewritten as follows:
\begin{eqnarray*}
  \pi(A|\sigma^2)
  &\propto&
  \int\cdots\int
  \prod_{(i_1,i_2)\in\mathcal{E}}
  \frac{1}{\sqrt{2\pi\sigma^2\tau_{i_1i_2}^2}}
  \exp\left\{
  -\frac{\sum_{j=1}^p(a_{i_1j}-a_{i_2j})^2}{2\sigma^2\tau_{i_1i_2}^2}
  \right\}
  \\
  &&\quad\quad\quad\quad\quad\quad
  \times
  \prod_{(i_1,i_2)\in\mathcal{E}}
  \frac{\lambda^2_1w_{i_1i_2}^2}{2}
  \exp\left\{
  -\frac{\lambda^2_1w_{i_1i_2}^2}{2}\tau^2_{i_1i_2}
  \right\}
  \\
  &&\quad\quad\quad\quad
  \times
  \prod_{j=1}^p
  \frac{1}{\sqrt{2\pi\sigma^2\widetilde{\tau}_j^2}}
  \exp\left\{
  -\frac{\sum_{i=1}^n a_{ij}^2}{2\sigma^2\widetilde{\tau}_j^2}
  \right\}
  \\
  &&\quad\quad\quad\quad\quad\quad
  \times
  \prod_{j=1}^p
  \frac{\lambda^2_2 u_j^2}{2}
  \exp\left\{
  -\frac{\lambda^2_2 u_j^2}{2}\widetilde{\tau}_j^2
  \right\}
  \prod_{(i_1,i_2)\in\mathcal{E}}d\tau^2_{i_1i_2}
  \prod_{j=1}^pd\widetilde{\tau}_j^2
  .
\end{eqnarray*}
This representation is based on the following hierarchical representation of the Laplace distribution:
\[
\frac{a}{2} \exp \{-a|z|\}
=
\int_{0}^{\infty} \frac{1}{\sqrt{2 \pi s}}
\exp \left\{-\frac{z^{2}}{2 s}\right\}
\frac{a^{2}}{2}
\exp \left\{-\frac{a^{2}}{2} s\right\} d s
.
\]
For details, we refer the reader to \citet{andrews1974scale}.

From this relationship, we assume the following priors:
\begin{eqnarray*}
  \pi\left(A | \{\tau_{i_1,i_2}^{2}\}, \{\widetilde{\tau}_j^{2}\}, \sigma^{2} \right)
  &\propto&
  \prod_{(i_1,i_2)\in\mathcal{E}}
  \frac{1}{\sqrt{\sigma^2\tau_{i_1i_2}^2}}
  \exp\left\{
  -\frac{\sum_{j=1}^p(a_{i_1j}-a_{i_2j})^2}{2\sigma^2\tau_{i_1i_2}^2}
  \right\}
  \\
  &&\qquad\quad\times
  \prod_{j=1}^p
  \frac{1}{\sqrt{\sigma^2\widetilde{\tau}_j^2}}
  \exp\left\{
  -\frac{\sum_{i=1}^n a_{ij}^2}{2\sigma^2\widetilde{\tau}_j^2}
  \right\}
  ,
  \\
  \pi\left(\tau_{i_1i_2}^{2}\right)
  &\propto&
  \frac{\lambda^2_1w_{i_1i_2}^2}{2}
  \exp\left\{
  -\frac{\lambda^2_1w_{i_1i_2}^2}{2}\tau^2_{i_1i_2}
  \right\}
  ,
  \\
  \pi\left(\widetilde{\tau}_j^2\right)
  &\propto&
  \frac{\lambda^2_2 u_j^2}{2}
  \exp\left\{
  -\frac{\lambda^2_2 u_j^2}{2}\widetilde{\tau}_j^2
  \right\}
  .
\end{eqnarray*}
These priors enable us to carry out Bayesian estimation using Gibbs sampling.
The details of the sampling procedure are described in Appendix \ref{gibbs_bscvc}.

\subsection{Unimodality of joint posterior distribution}

In Bayesian modeling, theoretical and computational problems arise when there exist multiple posterior modes.
Theoretically, it is doubtful whether a single posterior mean, median, or mode will appropriately summarize the bimodal posterior distribution.
The convergence speed of Gibbs sampling presents a computational problem, in that, although it is possible to perform Gibbs sampling, the convergence is too slow in practice.

\citet{Park_2008} showed that the joint posterior distribution has a single peak in Lasso-type Bayes sparse modeling.
We will demonstrate that the joint posterior distribution of (\ref{bscvc_prior_orig}) is unimodal.
Specifically, similar to \citet{Park_2008}, we will use a continuous transformation with a continuous inverse to show the unimodality of the logarithmic concave density.

The logarithm of the posterior (\ref{bscvc_prior_orig}) is
\begin{eqnarray}
  \log
  \pi(A, \sigma^2 | X)
  &=&
  \log \pi(\sigma^2)
  -\frac{pn + \#\mathcal{E} + p}{2}
  \log(\sigma^2)
  -\frac{1}{2\sigma^2}
  \sum_{i=1}^n\|\vector{x}_i-\vector{a}_i\|^2_2
  \nonumber \\&&
  -\frac{\lambda_1}{\sigma}
  \sum_{(i_1,i_2)\in\mathcal{E}}
  w_{i_1,i_2}
  \|\vector{a}_{i_1}-\vector{a}_{i_2}\|_2
  \nonumber \\&&
  -\frac{\lambda_2}{\sigma}
  \sum_{j=1}^p
  u_j
  \|\vector{a}_{\cdot j}\|_2
  +\mbox{const}
  .
  \label{log_posterior}
\end{eqnarray}
Consider the transformation defined by
\[
\Phi \leftrightarrow A/\sqrt{\sigma^2},\quad
\rho \leftrightarrow 1/\sqrt{\sigma^2},
\]
which is continuous when $0 < \sigma^2 < \infty$.
We define $\Phi=(\vector{\phi}_1,\cdots,\vector{\phi}_n)^T=(
\vector{\phi}_{\cdot 1},\cdots,\vector{\phi}_{\cdot p})$.
The log posterior (\ref{log_posterior}) is transformed by performing variable conversion in the form
\begin{eqnarray}
  &&
  \log \pi(1/\rho^2)
  +(pn+\#\mathcal{E} + p)
  \log(\rho)
  -\frac{1}{2}
  \sum_{i=1}^n\|\rho\vector{x}_i-\vector{\phi}_i\|^2_2
  \nonumber \\&&
  -\lambda_1
  \sum_{(i_1,i_2)\in\mathcal{E}}
  w_{i_1,i_2}
  \|\vector{\phi}_{i_1}-\vector{\phi}_{i_2}\|_2
  -\lambda_2
  \sum_{j=1}^p
  u_j
  \|\vector{\phi}_{\cdot j}\|_2
  +\mbox{const}
  \label{trans_log_posterior}
  .
\end{eqnarray}
The second and fifth terms are clearly concave in $(\Phi,\rho)$, and the third and fourth terms are a concave surface in $(\Phi,\rho)$.
Therefore, if $\log\pi(\cdot)$, which is the logarithm of the prior for $\sigma^2$, is concave, then (\ref{trans_log_posterior}) is concave.
Assuming a prior distribution, such as the inverse gamma distribution (\ref{inverse_gamma}) for $\sigma^2$, $\log\pi(\cdot)$ is a concave function.
Therefore, the entire log posterior distribution is concave.

\subsection{MAP estimate by weighted posterior means \label{weighted_posterior_means}}

In Bayesian sparse modeling, an unweighted posterior mean is often used as a substitute for MAP estimates, but the accuracy is not high and sometimes it is far from the MAP estimates.
As a result, we introduce the weighted posterior mean in this section.

We define a vector $\vector{\theta}$ containing all the parameters as follows:
\begin{eqnarray*}
\vector{\theta}
&=&
(\vector{\theta}_1,\cdots,\vector{\theta}_{2n+2})
\\
&=&
(
\vector{a}_1,\cdots,\vector{a}_n,
\vector{\tau}_1,\cdots,\vector{\tau}_n,
\widetilde{\vector{\tau}},
\sigma^2
)
,
\end{eqnarray*}
where
$\vector{\tau}_i=(\tau_{i1}, \cdots, \tau_{in})$ and
$\widetilde{\vector{\tau}}=(\widetilde{\tau}_{1}, \cdots, \widetilde{\tau}_{p})$.
For example, $\vector{\theta}_1=\vector{a}_1$ and $\vector{\theta}_{n+1}=\vector{\tau}_1$.
In addition, we assume the parameter vector corresponding to the $b$-th MCMC sample is $\vector{\theta}^{(b)}=(\vector{\theta}_1^{(b)},\cdots,\vector{\theta}_{2n+2}^{(b)})
$,
where the range of $b$ is from $1$ to $B$.

We introduce weights corresponding to the $b$-th MCMC sample as follows:
\begin{eqnarray*}
  \widetilde{w}_{(\vector{\theta}_l, b)}
  &=&
  L(X|\hat{\vector{\theta}}^{(b)}_{l})
  \pi(\hat{\vector{\theta}}^{(b)}_{l})
  ,
\end{eqnarray*}
where $L(X|\vector{\theta})$ is the likelihood function, $\pi(\vector{\theta})$ is the prior,
\[
\hat{\vector{\theta}}^{(b)}_{l} =
\{\hat{\vector{\theta}}_1,\cdots,\hat{\vector{\theta}}_{l-1},\vector{\theta}_{l}^{(b)},\hat{\vector{\theta}}_{l+1},\cdots,\hat{\vector{\theta}}_{2n+2}\},
\]
and $\hat{\vector{\theta}}_{l'}$ is an estimate of $\vector{\theta}_{l'}$.
It can be seen that this weight corresponds to the value of the posterior probability according to Bayes' theorem.
This weight was also used in the sparsified algorithm proposed by \citet{shimamura2019bayesian}.

Using this weight, we obtain the posterior average as follows:
\[
  \hat{\vector{\theta}}_l = \sum_{b=1}^B
  w_{(\vector{\theta}_l, b)}
  \vector{\theta}_l^{(b)}
  ,
\]
where $w_{(\vector{\theta}_l, b)}=\widetilde{w}_{(\vector{\theta}_l, b)}/\sum_{b'=1}^B\widetilde{w}_{(\vector{\theta}_l, b')}$.
Therefore, we adopt $\hat{\vector{\theta}}_l$ as an estimate of $\vector{\theta}_l$.
The performance of this estimate is examined by numerical studies in Section \ref{half_moons}.

\section{Bayesian sparse convex clustering via global-local (GL) shrinkage priors}

\citet{polson2010shrink} proposed a GL shrinkage prior distribution.
Generally speaking, when we use the Laplace prior distribution, it is necessary to pay attention to how to handle contraction for irrelevant parameters and robustness against relevant parameters.
The important features of the GL shrinkage prior distribution are that it has a peak at the origin and heavy tails.
These features make it possible to handle shrinkage of all variables, and the individual variables shrinkage estimated to be zero.
Therefore, irrelevant parameters are sparsified, and relevant ones are robustly estimated.
The penalty for sparse convex clustering has similar characteristics.
Specifically, it is weighted on individual and feature quantities.
This weighted penalty is one of the key factors for improving accuracy.
However, this penalty has the problem that it is highly dependent on the data.
By using the GL prior distribution, it is possible to properly control this the dependency by using the Bayesian approach.

\citet{polson2010shrink} formulated the GL scale mixtures of normal distributions for vector $\vector{a} = (a_1,\cdots,a_p)$ as follows:
\begin{eqnarray*}
  a_j | \nu^2, \tau_j^2 &\sim& \mbox{N}(0, \nu^2\tau_j^2)
  ,
  \\
  \tau_j^2 &\sim& \pi(\tau_j^2)
  ,
  \\
  \nu^2 &\sim& \pi(\nu^2)
  .
\end{eqnarray*}
Each $\tau_j^2\ (>0)$ is called a local shrinkage parameter and $\nu\ (>0)$ is called a global shrinkage parameter.
This leads to efficient Gibbs sampling based on block updating of parameters.

We need to specify the priors $\pi (\tau_j^2)$ and $\pi (\nu^2)$. In the next subsections, we provide some concrete formulations for $\pi (\tau_j^2)$ and $\pi (\nu^2)$.

\subsection{NEG prior distribution}

\citet{griffin2005alternative} proposed using an NEG distribution as an alternative to a Laplace distribution for the prior distribution of regression coefficients.
By using an NEG distribution, we can perform more flexible sparse modeling than with a Laplace distribution.

The NEG density function is given by
\begin{eqnarray}
\mbox{NEG}(\theta|\lambda,\gamma)
=
\kappa
\exp\left\{\frac{\theta^2}{4\gamma^2}\right\}
D_{-2\lambda-1}\left(\frac{|\theta|}{\gamma}\right)
,
\label{NEG_dist}
\end{eqnarray}
where
$\kappa = \displaystyle (2^\lambda\lambda)/(\gamma\sqrt{\pi}){\rm \Gamma}(\lambda+1/2)$ is a normalization constant, $D_{-2\lambda-1}$ is a parabolic cylinder function, and $\lambda\ (>0)$ and $\gamma\ (>0)$ are hyperparameters that control the sparsity of $\theta$.
The parabolic cylinder function is a solution of a second-order linear ordinary differential equation and its integral representation is given by
\begin{eqnarray*}
D_{-2\lambda-1}\left(\frac{|\theta|}{\gamma}\right)
=
\frac{1}{\rm \Gamma(2\lambda+1)}
\exp\left\{-\frac{\theta^2}{4\gamma^2}\right\}
\int_0^\infty
w^{2\lambda}
\exp\left\{
-\frac{1}{2}w^2-\frac{|\theta|}{\gamma}w
\right\}
dw
.
\end{eqnarray*}
The NEG density function can be expressed as hierarchical representation
\begin{eqnarray*}
  &&
  \mbox{NEG}\left(\theta|\lambda,\gamma\right)
  \\
  &=&
  \int\int
  \frac{1}{\sqrt{2\pi\tau^2}}
  \exp\left\{-\frac{\theta^2}{2\tau^2}\right\}
  \psi
  \exp\left\{-\psi\tau^2\right\}
  \frac{(\gamma^2)^\lambda}{\Gamma(\lambda)}
  \psi^{\lambda-1}
  \exp\left\{-\gamma^2\psi\right\}
  d\tau^2d\psi
  \\
  &=&
  \int\int
  \mbox{N}(\theta|0,\tau^2)
  \mbox{Exp}(\tau^2|\psi)
  \mbox{Ga}(\psi|\lambda,\gamma^2)
  d\tau^2d\psi
  ,
\end{eqnarray*}
where $\mbox{Exp}(\cdot|\mu)$ is the exponential distribution and $\mbox{Ga}(\cdot|k, \lambda)$ is a gamma distribution.
Therefore, the prior distribution of each parameter is as follows:
\begin{eqnarray*}
  \theta|\tau^2
  &\sim&
  \mbox{N}(\theta|0,\tau^2)
  ,\\
  \tau^2
  &\sim&
  \mbox{Exp}(\tau^2|\psi)
  ,\\
  \psi
  &\sim&
  \mbox{Ga}(\psi|\lambda,\gamma^2)
  .
\end{eqnarray*}

Using the NEG distribution on the feature matrix $A$, we propose the following prior:
\begin{eqnarray*}
  \pi(A|\sigma^2)
  &\propto&
  (\sigma^2)^{-(\#\mathcal{E}+p)/2}
  \prod_{(i_1,i_2)\in\mathcal{E}}
  \mbox{NEG}\left(
  \frac{1}{2\sigma}
  \|\vector{a_{i_1}}-\vector{a_{i_2}}\|_2
  \ \Big|\
  \lambda_1,\gamma_1
  \right)
  \\
  &&\quad\quad\times
  \prod_{j=1}^p
  \mbox{NEG}\left(
  \frac{1}{2\sigma}
  \|\vector{a_{\cdot j}}\|_2
  \ \Big|\
  \lambda_2,\gamma_2
  \right)
  .
\end{eqnarray*}
By using the hierarchical representation of the NEG distribution, the prior distribution $\pi(A|\sigma^2)$ is decomposed into
\begin{eqnarray*}
  \pi(A|\sigma^2)
  &\propto&
  \int\cdots\int
  \prod_{(i_1,i_2)\in\mathcal{E}}
  (\sigma^2\tau_{i_1i_2}^2)^{-1/2}
  \exp\left\{
     -\frac{1}{2\sigma^2\tau_{i_1i_2}^2}
     \|\vector{a}_{i_1}-\vector{a}_{i_2}\|_2^2
  \right\}
  \\
  &&\quad\times
  \prod_{(i_1,i_2)\in\mathcal{E}}
  \psi_{i_1i_2}\exp\{-\psi_{i_1i_2}\tau_{i_1i_2}^2\}
  \prod_{(i_1,i_2)\in\mathcal{E}}
  \frac{(\gamma_1^2)^{\lambda_1}}{\Gamma(\lambda_1)}
  \psi_{i_1i_2}^{\lambda_1-1}\exp\{-\gamma_1^2\psi_{i_1i_2}\}
  \\
  &&\quad\times
  \prod_{j=1}^p
  (\sigma^2\widetilde{\tau}_j^2)^{-1/2}
  \exp\left\{
    -\frac{1}{2\sigma^2\widetilde{\tau}_j^2}
    \|\vector{a}_{\cdot j}\|_2^2
  \right\}
  \\
  &&\quad\times
  \prod_{j=1}^p
  \widetilde{\psi}_j\exp\{-\widetilde{\psi}_j\widetilde{\tau}_j^2\}
  \prod_{j=1}^p
  \frac{(\gamma_2^2)^{\lambda_2}}{\Gamma(\lambda_2)}
  \widetilde{\psi}_j^{\lambda_2-1}\exp\{-\gamma_2^2\widetilde{\psi}_j\}
  \\
  &&\quad\times
  \prod_{(i_1,i_2)\in\mathcal{E}}d\psi_{i_1i_2}
  \prod_{(i_1,i_2)\in\mathcal{E}}d\tau_{i_1i_2}^2
  \prod_{j=1}^pd\widetilde{\psi}_j
  \prod_{j=1}^pd\widetilde{\tau}_j^2
  .
\end{eqnarray*}
This result allows us to develop a Gibbs sampling algorithm for Bayesian sparse convex clustering with the NEG prior distribution.
The details of the algorithm are given in Appendix \ref{negscvc_gibbs}.
\subsection{Horseshoe prior distribution}

The horseshoe density function \citep{carvalho2010horseshoe} is given by
\begin{eqnarray*}
  \mbox{Hor}(\vector{\theta}|\nu)
  &\propto&
  \int\cdots\int
  \prod_{j=1}^p
  \left\{
  p(\theta_j|\tau_j^2,\nu)
  p(\tau_j^2)
  \right\}
  \prod_{j=1}^pd\tau_j
  \\
  &\propto&
  \int\cdots\int
  \prod_{j=1}^p
  \left\{
  p(\theta_j|\tau_j^2,\nu)
  p(\tau_j^2|\psi_j)
  p(\psi_j)
  \right\}
  \prod_{j=1}^p(d\tau_jd\psi_j)
  .
\end{eqnarray*}
The prior distribution of each parameter is as follows:
\begin{eqnarray*}
  \theta_j|\tau_j^2,\nu^2
  &\sim&
  \mbox{N}(0, \tau_j^2\nu^2)
  ,\\
  \tau_j^2
  &\sim&
  \mbox{C}^+(0,1)
  ,\\
  \tau_j^2|\psi_j
  &\sim&
  \mbox{IG}(1/2,1/\psi_j)
  ,\\
  \psi_j
  &\sim&
  \mbox{IG}(1/2,1)
  .
\end{eqnarray*}
Here $\nu\ (>0)$ is a hyperparameter that controls the sparsity of the $\theta_j$'s, and $\mbox{C}^+(x_0, \gamma)$ is the half Cauchy distribution on the positive reals, where
$x_0$ is a location parameter and $\gamma$ is a scale parameter.
A smaller value of hyperparameter $\nu$ corresponds to a higher, number of parameters $\{\theta_j\}$ being estimated to be zero.

Using the horseshoe distribution on the feature matrix $A$, we propose the following prior:
\begin{eqnarray}
  \pi(A|\sigma^2)
  &\propto&
  (\sigma^2)^{-(\#\mathcal{E}+p)/2}
  \mbox{Hor}\left(
  \frac{1}{2\sigma}
  \vector{\mathrm{a}}
  \ \Big|\
  \nu_1
  \right)
   \nonumber\\ &&\qquad\times
  \prod_{j=1}^p
  \mbox{NEG}\left(
  \frac{1}{2\sigma}
  \|\vector{a}_{\cdot j}\|_2
  \ \Big|\
  \lambda_2, \gamma_2
  \right)
  ,
  \label{bscvc_horneg}
\end{eqnarray}
where
$\vector{\mathrm{a}}
 = \left({\|\vector{a}_{i_1}-\vector{a}_{i_2}\|_2;\ (i_1,i_2)\in\mathcal{E}} \right)$.
Note that this prior distribution consists of the horseshoe distribution and the NEG distribution.
The prior distribution can also be constructed using only the horseshoe distribution.
However, as a result of the numerical experiment, it did not work well.
Therefore, we adopt the NEG distribution for the prior that induces variable selection.

By using the hierarchical representation of the horseshoe distribution, the prior distribution $\pi(A|\sigma^2)$ is obtained as follows:
\begin{eqnarray*}
  \pi(A|\sigma^2)
  &\propto&
  \int\cdots\int
  \prod_{(i_1,i_2)\in\mathcal{E}}(\sigma^2\tau_{i_1i_2}^2\nu^2_1)^{-1/2}
  \exp\left\{
    -\frac{1}{2\sigma^2\tau_{i_1i_2}^2\nu^2_1}
    \|\vector{a}_{i_1}-\vector{a}_{i_2}\|_2^2
  \right\}
  \\
  &&\quad\quad\times
  \prod_{(i_1,i_2)\in\mathcal{E}}
  \psi_{i_1i_2}^{-1/2}(\tau_{i_1i_2}^2)^{-1/2-1}
  \exp\left\{
    -\frac{1}{\psi_{i_1i_2}\tau_{i_1i_2}^2}
  \right\}
  \\
  &&\quad\quad\times
  \prod_{(i_1,i_2)\in\mathcal{E}}
  \psi_{i_1i_2}^{-1/2-1}
  \exp\left\{
    -\frac{1}{\psi_{i_1i_2}}
  \right\}
  \\
  &&\quad\quad\times
  \prod_{j=1}^p
  (\sigma^2\widetilde{\tau}_j^2)^{-1/2}
  \exp\left\{
    -\frac{1}{2\sigma^2\widetilde{\tau}_j^2}
    \|\vector{a}_{\cdot j}\|_2^2
  \right\}
  \\
  &&\quad\quad\times
  \prod_{j=1}^p
  \widetilde{\psi}_j\exp\{-\widetilde{\psi}_j\widetilde{\tau}_j^2\}
  \prod_{j=1}^p
  \frac{(\gamma_2^2)^{\lambda_2}}{\Gamma(\lambda_2)}
  \widetilde{\psi}_j^{\lambda_2-1}\exp\{-\gamma_2^2\widetilde{\psi}_j\}
  \\
  &&\quad\quad\times
  \prod_{(i_1,i_2)\in\mathcal{E}} d\tau_{i_1i_2}
  \prod_{(i_1,i_2)\in\mathcal{E}} d\psi_{i_1i_2}
  \prod_{j=1}^pd\widetilde{\psi}_j
  \prod_{j=1}^pd\widetilde{\tau}_j^2
  .
\end{eqnarray*}
Then we can estimate the posterior distribution by Gibbs sampling.
The details of the algorithm are given in Appendix \ref{horscvc_gibbs}.

\subsection{Dirichlet--Laplace prior distribution}

The Dirichlet--Laplace prior was proposed to provide simple sufficient conditions for posterior consistency \citep{bhattacharya2015dirichlet}.
It is known that a Bayesian regression model with this prior distribution has asymptotic posterior consistency with respect to variable selection.
Also, we can obtain joint posterior distributions for a Bayesian regression model when we employ this prior.
The latter is advantageous because most prior distributions induce a marginal posterior distribution rather than a joint posterior distribution, which has more information in general.

The Dirichlet--Laplace density function is given by
\begin{eqnarray*}
  \mbox{DL}(\vector{\theta}|\alpha)
  &\propto&
  \int\cdots\int
  \prod_{j=1}^p
  \left\{
    p(\theta_j|\tau_j,\nu)
  \right\}
  p(\vector{\tau}|\alpha)
  p(\nu)
  \prod_{j=1}^p(d\tau_j)
  d\nu
  \\
  &\propto&
  \int\cdots\int
  \prod_{j=1}^p
  \left\{
    p(\theta_j|\psi_j,\tau_j^2,\nu^2)p(\psi_j)
  \right\}
  p(\vector{\tau}|\alpha)
  p(\nu)
  \prod_{j=1}^p(d\tau_j\psi_j)
  d\nu
  ,
\end{eqnarray*}
where $\vector{\tau} = (\tau_1,\cdots,\tau_p)^T$.
The prior distribution of each parameter is
\begin{eqnarray*}
  \theta_j|\tau_j,\nu
  &\sim&
  \mbox{Laplace}(1/\tau_j\nu),
  \\
  \theta_j|\tau_j,\psi_j,\nu
  &\sim&
  \mbox{N}(0, \psi_j\tau_j^2\nu^2),
  \\
  \vector{\tau}
  &\sim&
  \mbox{Dir}(\alpha,\cdots,\alpha),
  \\
  \psi_j
  &\sim&
  \mbox{Exp}(1/2),
  \\
  \nu
  &\sim&
  \mbox{Ga}(p\alpha,1/2),
\end{eqnarray*}
where $\alpha\ (>0)$ is a hyperparameter that controls the sparsity of the $\theta_j$'s and $\mbox{Dir}(\alpha,\cdots,\alpha)$ is a Dirichlet distribution.
The Dirichlet distribution random variables sum to one, and have mean $\mbox{E}[\tau_j]=1/p$ and variance $\mbox{Var}(\tau_j) = (p-1)/\{p^2(p\alpha+1)\}$.
When $\alpha$ is small, most of the parameters $\{\tau_j\}$ are close to zero, whereas the remaining parameters are close to one.
If $\{\tau_j\}$ is close to zero, $\{\theta_j\}$ is also close to zero.

Using the Dirichlet--Laplace distribution on the feature matrix $A$, we propose the following prior:
\begin{eqnarray*}
  \pi(A|\sigma^2)
  &\propto&
  (\sigma^2)^{-(\#\mathcal{E}+p)/2}
  \mbox{DL}\left(
  \frac{1}{2\sigma}
  \vector{\mathrm{a}}
  \ \Big|\
  \alpha_1
  \right)
  \prod_{j=1}^p
  \mbox{NEG}\left(
  \frac{1}{2\sigma}
  \|\vector{a}_{\cdot j}\|_2
  \ \Big|\
  \lambda_2, \gamma_2
  \right).
\end{eqnarray*}
Similar with reasons as in the prior (\ref{bscvc_horneg}), this prior distribution consists of the Dirichlet-Laplace distribution and the NEG distribution
By using a hierarchical representation of the Dirichlet--Laplace distribution, the prior distribution $\pi(A|\sigma^2)$ is obtained as follows:
\begin{eqnarray*}
  \pi(A|\sigma^2)
  &\propto&
  \int\cdots\int
    \prod_{(i_1,i_2)\in\mathcal{E}}
    (\sigma^2\psi_{i_1i_2}\tau_{i_1i_2}^2\nu^2)^{-1/2}
    \exp\left\{
      -\frac{1}{2\sigma^2\psi_{i_1i_2}\tau_{i_1i_2}^2\nu^2}
      \|\vector{a}_{i_1}-\vector{a}_{i_2}\|_2^2
    \right\}
  \\
  &&\quad\quad\times
    \prod_{(i_1,i_2)\in\mathcal{E}}
    \frac{1}{2}\exp\left\{
      -\frac{1}{2}\psi_{i_1i_2}
    \right\}
    \prod_{(i_1,i_2)\in\mathcal{E}} \tau_{i_1i_2}^{\alpha_1-1}
  \\
  &&\quad\quad\times
    \nu^{\alpha_1\#\mathcal{E}-1}
    \exp\left\{-\frac{1}{2}\nu\right\}
  \\
  &&\quad\quad\times
    \prod_{j=1}^p
    (\sigma^2\widetilde{\tau}_j^2)^{-1/2}
    \exp\left\{
      -\frac{1}{2\sigma^2\widetilde{\tau}_j^2}
      \|\vector{a}_{\cdot j}\|_2^2
    \right\}
  \\
  &&\quad\quad\times
    \prod_{j=1}^p
    \widetilde{\psi}_j\exp\{-\widetilde{\psi}_j\widetilde{\tau}_j^2\}
    \prod_{j=1}^p
    \frac{(\gamma_2^2)^{\lambda_2}}{\Gamma(\lambda_2)}
    \widetilde{\psi}_j^{\lambda_2-1}\exp\{-\gamma_2^2\widetilde{\psi}_j\}
  \\
  &&\quad\quad\times
  d\nu
  \prod_{(i_1,i_2)\in\mathcal{E}}d\tau_{i_1i_2}
  \prod_{(i_1,i_2)\in\mathcal{E}}d\psi_{i_1i_2}
  \prod_{j=1}^pd\widetilde{\psi}_j
  \prod_{j=1}^pd\widetilde{\tau}_j^2
  .
\end{eqnarray*}
Then we can estimate the posterior distribution by using Gibbs sampling.
The details of the algorithm are given in Appendix \ref{dlscvc_gibbs}.

\section{Artificial data analysis \label{simu_study}}

In this section, we describe numerical studies to evaluate the performance of the proposed methods using artificial data.
First, clustering performance was evaluated by an illustrative example that includes no irrelevant features.
Next, we evaluated the accuracy of the sparsity by performing simulations using data containing irrelevant features.

\subsection{Illustrative example \label{half_moons}}

\begin{figure}
  \begin{center}
    \includegraphics[bb=0 0 960 960 ,width=90mm]{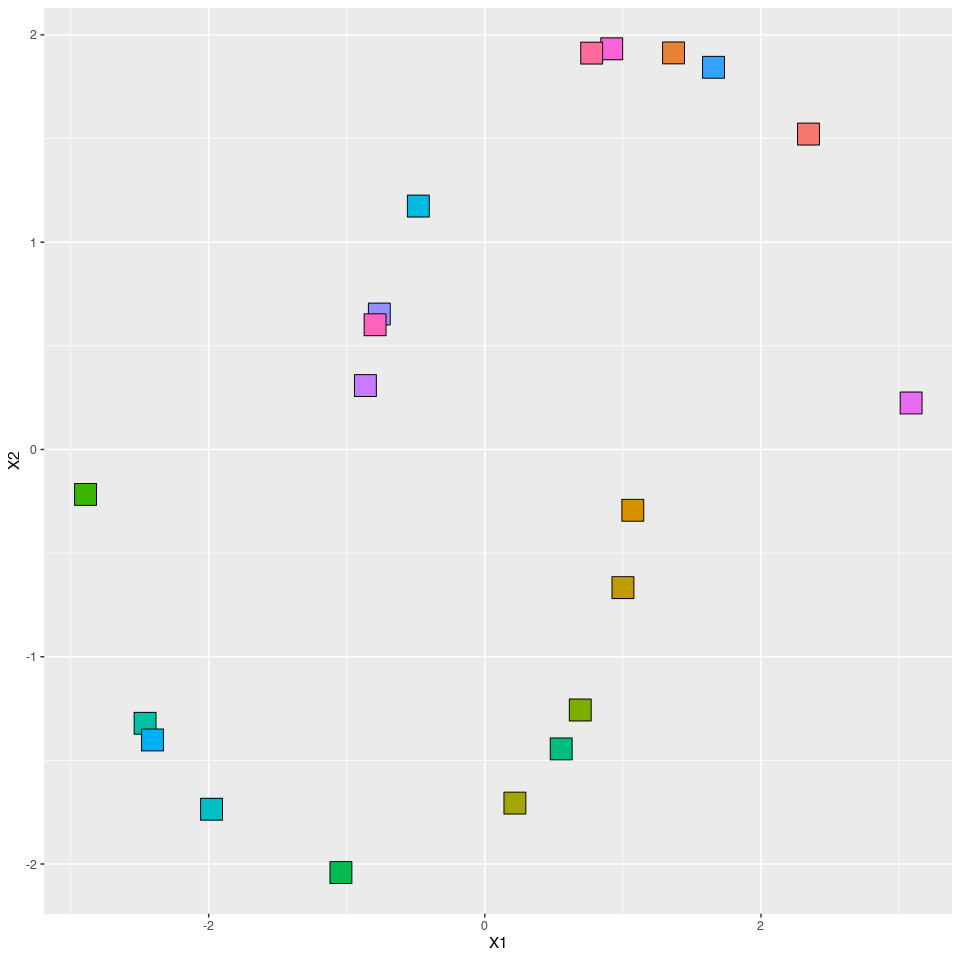}
  \end{center}
  \caption{
  Two interlocking half-moons with $n=50$ observations.
  }
  \label{halfmoons_data}
\end{figure}

We demonstrated our proposed methods with artificial data.
The data were generated according to two interlocking half-moons with $n=50$ observations, $K=2$ clusters, and $p=2$ features.
Figure \ref{halfmoons_data} shows one example of two interlocking half-moons.
In this setting, we did not perform sparse estimation.
The cluster formation was considered by comparing the cluster paths of each method.

For each generated dataset, the estimates were obtained by using 50,000 iterations of a Gibbs sampler.
Candidates of the hyperparameters were set based on
\[
\lambda_{\min}
\exp\{
  (\log\lambda_{\max} - \log\lambda_{\min}) \cdot (i/m)
\}
\]
for $i=1,\cdots,m$.
For the hyperparameter $\lambda$ in Bayesian convex clustering with a Laplace prior distribution (Bscvc),
we set $m = 50$, $\lambda_{\mbox{min}} = 0.05$, and $\lambda_{\mbox{max}} = 90.0$.
In Bayesian convex clustering with an NEG prior distribution (Bnegscvc), we had hyperparameters $\lambda_1$ and $\gamma_1$.
For hyperparameter $\lambda_1$, we set $m = 30$, $\lambda_{\mbox{min}} = 0.0001$, and $\lambda_{\mbox{max}} = 2.75$.
For hyperparameter $\gamma_1$, we set $m = 2$, $\lambda_{\mbox{min}} = 0.4$, and $\lambda_{\mbox{max}} = 0.5$.
The weighted posterior means introduced in Section \ref{weighted_posterior_means} were used for Bscvc and Bnegscvc estimates.

\begin{figure}[h]
  \begin{minipage}[b]{0.49\linewidth}
    \centering
    \includegraphics[bb=0 0 960 960 ,width=70mm, clip]{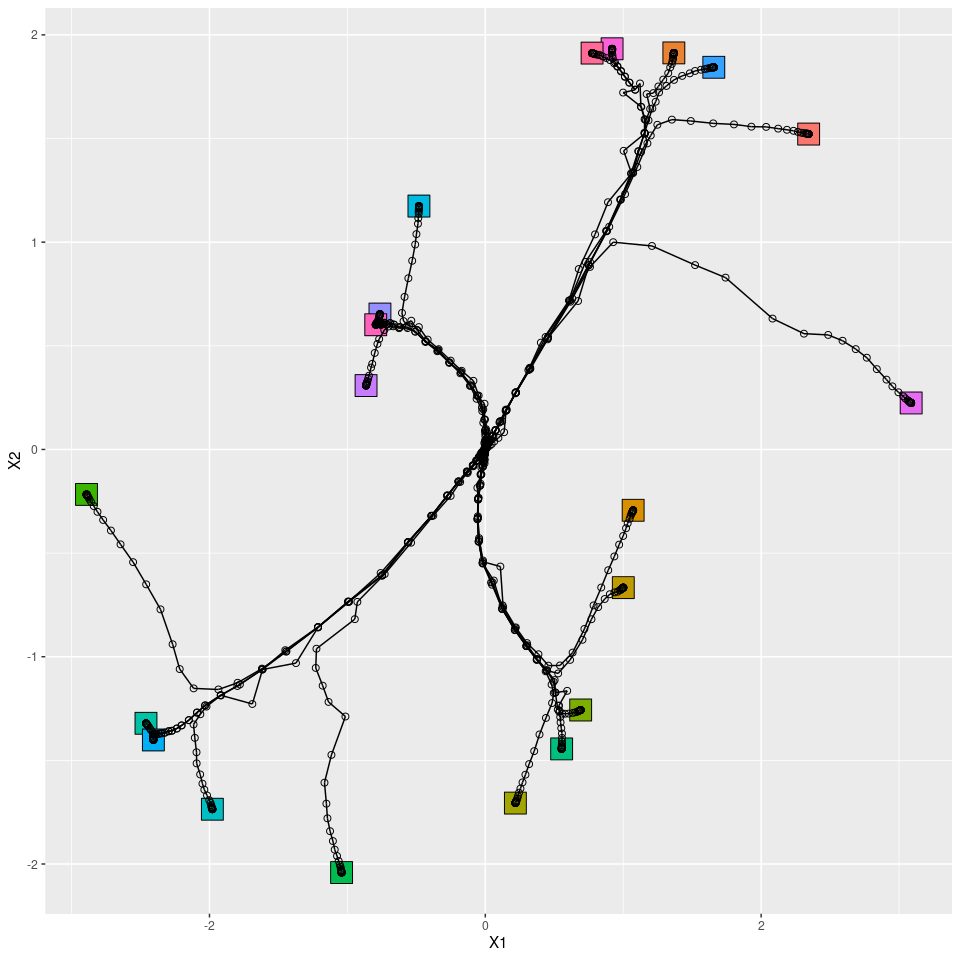}
    \\ \hspace{-1.6cm}
    (a)
  \end{minipage}
  \begin{minipage}[b]{0.49\linewidth}
    \centering
    \includegraphics[bb=0 0 960 960 ,width=70mm, clip]{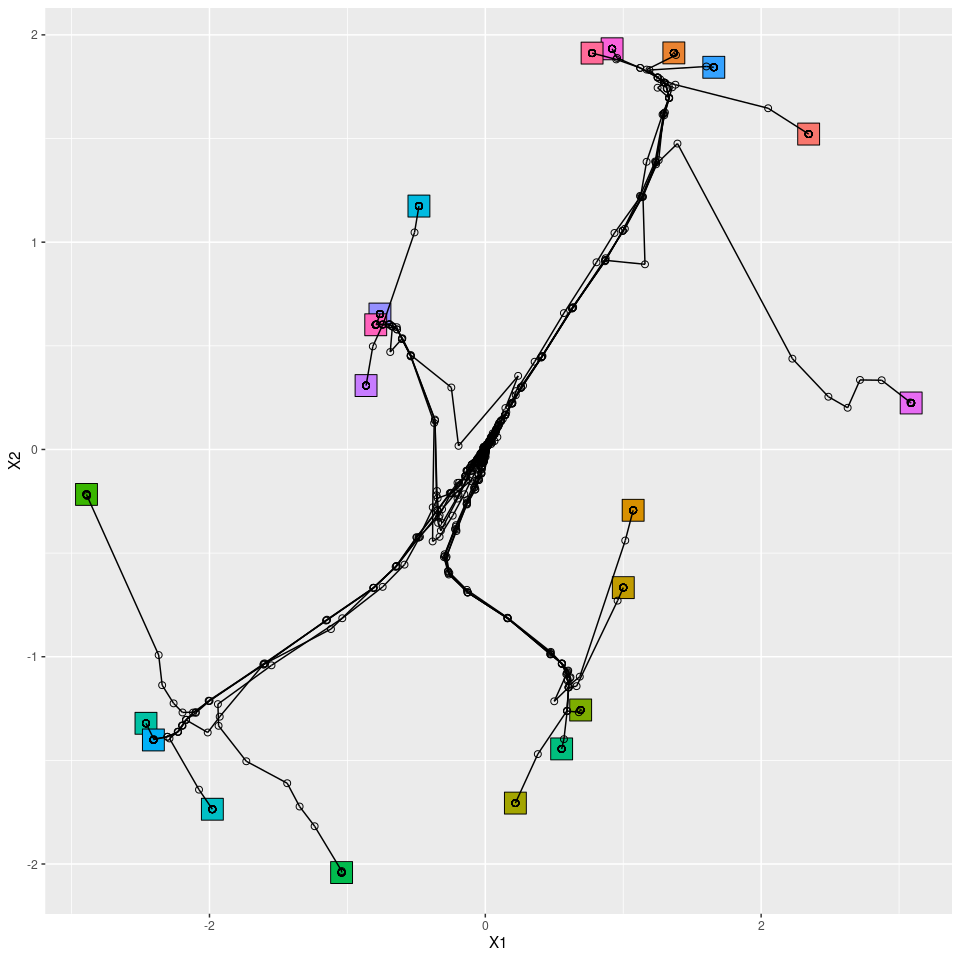}
    \\ \hspace{-1.6cm}
    (b)
  \end{minipage}
  \begin{minipage}[b]{0.49\linewidth}
    \centering
    \includegraphics[bb=0 0 960 960 ,width=70mm, clip]{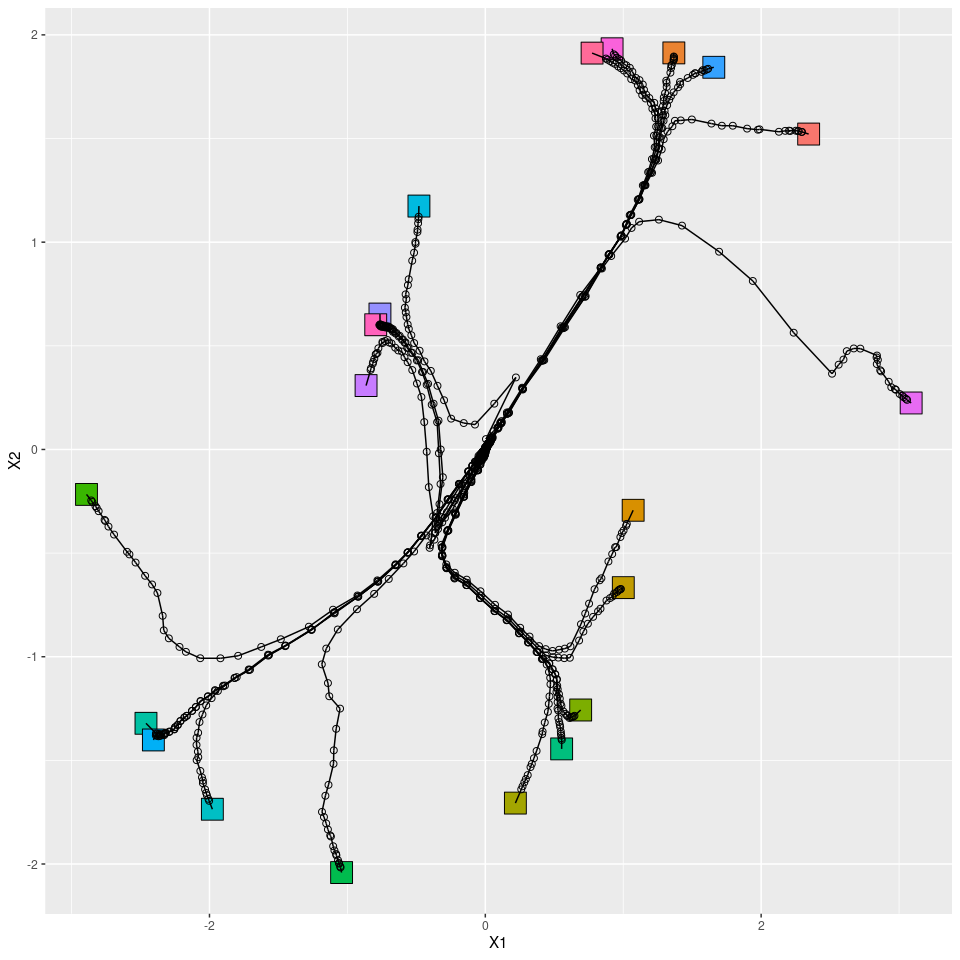}
    \\ \hspace{-1.6cm}
    (c)
  \end{minipage}
  \begin{minipage}[b]{0.49\linewidth}
    \centering
    \includegraphics[bb=0 0 960 960 ,width=70mm, clip]{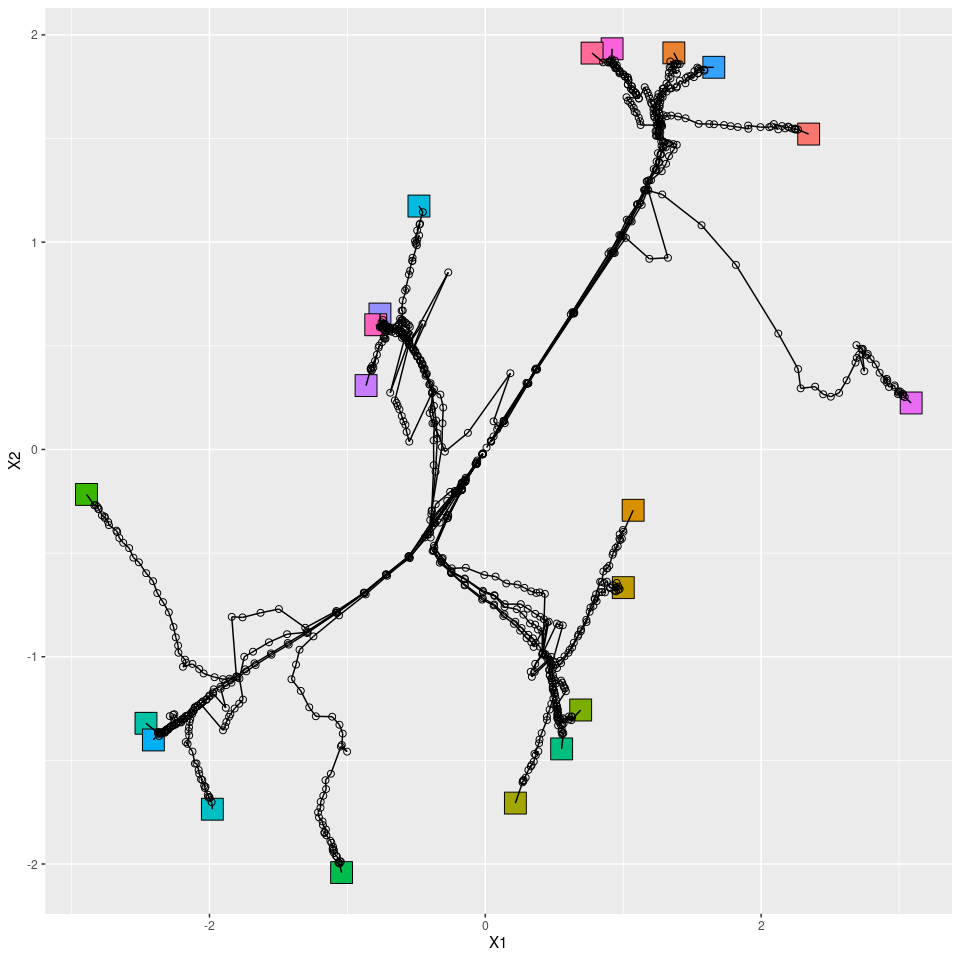}
    \\ \hspace{-1.6cm}
    (d)
  \end{minipage}

  \caption{
  Results for two interlocking half-moons.
  (a) Bscvc, (b) Bnegscvc, (c) Bhorscvc, (d) Bdlscvc.
  }
  \label{toyHalfMoonsResult}
\end{figure}

Figure \ref{toyHalfMoonsResult} shows the results.
The overall outline of cluster formation is the same for the all methods.
The order in which the samples form clusters is also the same.
If the distance between estimated feature values of different clusters does not decrease, the accuracy of cluster estimation will improve in convex clustering.
However, the distances between all features are small due to the effect of sparse regularization.
Scvc used weights to bring only features belonging to the same cluster closer.
Bnegscvc, Bhorscvc, and Bdlscvc used GL priors instead of weights.
For example, in the cluster path in Figure \ref{toyHalfMoonsResult}(b), the estimated feature values are merged at a position further from the origin than other methods.
This can be seen especially in the upper right and lower left of the figure.
This result shows that the close feature values were merged while the distances between the distant feature values were maintained.
This is a factor that improves the accuracy of Bnegscvc's clustering estimation.

\subsection{Simulation studies \label{half_moons_simu}}

We demonstrated our proposed methods with artificial data including irrelevant features.
First, we considered five settings.
Each data were generated according to two interlocking half-moons with  $n=10$, $20$, $50$, $100$ observations, $K=2$ clusters, and $p=20$, $40$ features.
The features consisted $p-2$  irrelevant features and 2 relevant features.
The irrelevant features were independently generated from $\mbox{N}(0, 0.5^2)$.
We considered three methods: sparse convex clustering (Scvc), Bscvc, and Bnegscvc.

As the estimation accuracy, we used the RAND index, which is a measure of correctness of cluster estimation.
The RAND index ranges between $0$ and $1$, with a higher value indicating better performance.
The RAND index is given by
\begin{eqnarray*}
\mbox{RAND}=\frac{a+b}{n(n-1)/2},
\end{eqnarray*}
where
\begin{eqnarray*}
a&=&
\sum_{k=1}^r\sum_{l=1}^s
\#\left\{
(\vector{x}_i,\vector{x}_j)
|
\vector{x}_i,\vector{x}_j \in \mathcal{C}_k^*,
\vector{x}_i,\vector{x}_j \in \widetilde{\mathcal{C}}_l;
i<j
\right\},
\\
b&=&
\sum_{k_1<k_2}\sum_{l_1<l_2}
\#\left\{
(\vector{x}_i,\vector{x}_j)
|
\vector{x}_i \in \mathcal{C}_{k_1}^*,
\vector{x}_j \in \mathcal{C}_{k_2}^*,
\vector{x}_i \in \widetilde{\mathcal{C}}_{l_1},
\vector{x}_j \in \widetilde{\mathcal{C}}_{l_2};
i<j
\right\}.
\end{eqnarray*}
Here $\mathcal{C}^*=\{\mathcal{C}_1^*,\cdots,\mathcal{C}_r^*\}$ is the true set of clusters and $\widetilde{\mathcal{C}}=\{\widetilde{\mathcal{C}}_1,\cdots,\widetilde{\mathcal{C}}_s\}$ is the estimated set of clusters.
In addition, we used the true negative rate (TNR) and the
true positive rate (TPR) for the accuracy of sparse estimation:
\begin{eqnarray*}
  \mbox{TNR}
  =
  \frac{\#\{j | \hat{\vector{a}}_j = \vector{0} \land \vector{a}_j^* = \vector{0}\}}{\#\{j | \vector{a}_j^* = \vector{0}\}}
  ,\quad
  \mbox{TPR}
  =
  \frac{\#\{j | \hat{\vector{a}}_j \neq \vector{0} \land \vector{a}_j^* \neq \vector{0}\}}{\#\{j | \vector{a}_j^* \neq \vector{0}\}}
  ,
\end{eqnarray*}
where, $\{\vector{a}_j^*|j=1,\cdots,p\}$ are the true feature vectors and $\{\hat{\vector{a}}_j|j=1,\cdots,p\}$ are the estimated feature vectors.
Estimated indicators are calculated $50$ times.
The settings of the iteration count and the hyperparameter candidate were the same as given in Section \ref{half_moons}.
To ensure fair comparisons, we used the results with hyperparameters that maximize the RAND index.

The simulation results were summarized in Table \ref{halfMoonsTable}.
Scvc provided the lower RANDs and TNRs than other methods in all settings. TPR was competitive among the all methods. Except for Scvc, Bnegscvc, Bhorscvc, and Bdlscvc were better than Bscvc in terms of RAND in almost all settings. From these experiments, we observed that the Bayesian convex clustering methods was superior to the conventional convex clustering method. In addition,  the Bayesian methods based on the GL priors relatively produced the higher RANDs than those based on the Laplace prior.

\begin{table}[htbp]
  \begin{center}
    \caption{Results for simulation study.}
    \label{halfMoonsTable}
    \begin{tabular}{ccccccc}
      \hline
      & \multicolumn{6}{c}{$n=10,\ p=20$}\\
      \cline{2-7}
               & RAND & (sd) & TNR & (sd) & TPR & (sd)  \\
      \hline
      Scvc		 & 0.58	& (0.22) & 0.67 & (0.37) & 0.98 & (0.10) \\
      Bscvc    & 0.70	& (0.23) & 0.98 & (0.05) & 1.00 & (0.00) \\
      Bnegscvc & 0.72	& (0.23) & 0.91 & (0.11) & 0.99 & (0.07) \\
      Bhorscvc & 0.78	& (0.20) & 0.98 & (0.04) & 1.00 & (0.00) \\
      Bdlscvc  & 0.88	& (0.17) & 0.99 & (0.02) & 0.99 & (0.07) \\
      \hline
      \\
      \hline
      & \multicolumn{6}{c}{$n=20,\ p=20$}\\
      \cline{2-7}
               & RAND & (sd) & TNR & (sd) & TPR & (sd)  \\
      \hline
      Scvc		 & 0.68	& (0.20) & 0.72 & (0.26) & 0.97 & (0.12) \\
      Bscvc    & 0.78	& (0.18) & 0.96 & (0.19) & 1.00 & (0.00) \\
      Bnegscvc & 0.82	& (0.16) & 0.92 & (0.16) & 1.00 & (0.00) \\
      Bhorscvc & 0.91	& (0.13) & 0.97 & (0.03) & 1.00 & (0.00) \\
      Bdlscvc  & 0.92	& (0.13) & 0.97 & (0.04) & 1.00 & (0.00) \\
      \hline
      \\
      \hline
      & \multicolumn{6}{c}{$n=20,\ p=40$}\\
      \cline{2-7}
               & RAND & (sd) & TNR & (sd) & TPR & (sd)  \\
      \hline
      Scvc		 & 0.62	& (0.22) & 0.78 & (0.23) & 0.97 & (0.16) \\
      Bscvc    & 0.68	& (0.21) & 0.95 & (0.17) & 1.00 & (0.00) \\
      Bnegscvc & 0.76	& (0.18) & 0.93 & (0.13) & 1.00 & (0.00) \\
      Bhorscvc & 0.85	& (0.18) & 0.99 & (0.02) & 0.99 & (0.07) \\
      Bdlscvc  & 0.84	& (0.19) & 0.99 & (0.02) & 1.00 & (0.00) \\
      \hline
      \\
      \hline
      & \multicolumn{6}{c}{$n=50,\ p=40$}\\
      \cline{2-7}
               & RAND & (sd) & TNR & (sd) & TPR & (sd)  \\
      \hline
      Scvc		 & 0.73	& (0.18) & 0.44 & (0.40) & 1.00 & (0.00) \\
      Bscvc    & 0.90	& (0.15) & 1.00 & (0.01) & 1.00 & (0.00) \\
      Bnegscvc & 0.99	& (0.11) & 0.99 & (0.04) & 1.00 & (0.00) \\
      Bhorscvc & 0.93	& (0.09) & 0.94 & (0.05) & 1.00 & (0.00) \\
      Bdlscvc  & 0.94	& (0.08) & 0.93 & (0.05) & 1.00 & (0.00) \\
      \hline
      \\
      \hline
      & \multicolumn{6}{c}{$n=100,\ p=40$}\\
      \cline{2-7}
               & RAND & (sd) & TNR & (sd) & TPR & (sd)  \\
      \hline
      Scvc		 & 0.79	& (0.18) & 0.74 & (0.32) & 1.00 & (0.00) \\
      Bscvc    & 0.99	& (0.02) & 0.90 & (0.07) & 1.00 & (0.00) \\
      Bnegscvc & 0.93	& (0.10) & 0.99 & (0.03) & 1.00 & (0.00) \\
      Bhorscvc & 0.97	& (0.03) & 0.83 & (0.09) & 1.00 & (0.00) \\
      Bdlscvc  & 0.96	& (0.04) & 0.84 & (0.08) & 1.00 & (0.00) \\
      \hline
    \end{tabular}
  \end{center}
\end{table}

\section{Application \label{application}}

We applied our proposed methods to a real dataset: the LIBRAS movement data from the Machine Learning Repository \citep{lichman2013uci}.
The LIBRAS movement dataset has 15 classes.
Each class was divided by type of hand movement.
There are 24 observations in each class, and each observation has 90 features consisting of hand movement coordinates.
In this numerical experiment, 5 classes were selected from among the 15 classes that were the same classes as those selected by \citet{wang2018sparse}.
Accuracies of each method were evaluated using the RAND index, the estimated number of clusters, and the number of selected features.
This is the same procedure as reported in \citet{wang2018sparse}.
As in Section \ref{half_moons_simu}, we used the results with hyperparameters that maximize the RAND index for comparisons.

The results are summarized in Table \ref{librasTable}.
For RAND, Bnegscvc was slightly higher than other methods. Also all the methods except for Bnegscvc provided the same RAND. Bnegscvc selected six clusters, while other methods selected five clusters. Although the true number of clusters is five, the inherent number of clusters might be six because the corresponding RAND is highest among all methods. Scvc, Bscvc, and Bnegscvc selected all features, while Bhorscvc and Bdlscvc selected some of features.
In other words, Bhorscvc and Bdlscvc could be sparsified without degrading the accuracy of cluster estimation.

\begin{table}[htbp]
\begin{center}
\caption{Application to LIBRAS movement dataset.}
\label{librasTable}
\begin{tabular}{ccccccc}
  \hline
   & RAND & Clusters & Selected features  \\
  \hline
  Scvc		 & 0.767	& 5 & 90 \\
  Bscvc    & 0.767	& 5 & 90 \\
  Bnegscvc & 0.774	& 6 & 90 \\
  Bhorscvc & 0.767	& 5 & 68 \\
  Bdlscvc  & 0.767	& 5 & 38 \\
  \hline
  \end{tabular}
  \end{center}
\end{table}

\section{Conclusion}

We proposed a Bayesian formulation of the sparse convex clustering.
Using the GL shrinkage prior distribution, we constructed a Bayesian model for various data with more flexible constraints than ordinary $L_1$-type convex clustering.
We overcame the problem that sparse convex clustering depends on weights in the regularization term.
Furthermore, we proposed a weighted posterior mean based on a posteriori probability to provide more accurate MAP estimation.

For the application described in Section \ref{application}, the computational time with our proposed methods was about 20 minutes for each hyperparameter.
Using the GL shrinkage prior increases the computational cost, and hence we need to balance the feasibility of the calculation with the accuracy of the estimation.
In our numerical experiment, the hyperparameters with the best accuracy were selected using the same method as reported in \citet{wang2018sparse}.
It would also be interesting to develop information criteria for selecting the hyperparameters.
We leave these topics as future work.

\appendix
\def\thesection{Appendix}

\section{Formulation of Gibbs sampling}
This appendix introduces a specific Gibbs sampling method for a Bayesian sparse convex clustering.

\def\thesection{\Alph{section}}

\subsection{Bayesian sparse convex clustering\label{gibbs_bscvc}}

The prior distribution is transformed as follows:
\begin{eqnarray*}
  &&
  \pi(A,\{\tau_{i_1i_2}\},\{\widetilde{\tau}_j\},\sigma^2|X)
  \\
  &&\quad\propto
  (2\pi\sigma^2)^{-pn/2}
  \exp\left\{
  -\frac{1}{2\sigma^2}
  \sum_{i=1}^n
  (\vector{x}_i-\vector{a}_i)^T(\vector{x}_i-\vector{a}_i)
  \right\}
  \\
  &&\quad\quad\times
  (\sigma^2)^{-(\#\mathcal{E}+p)/2}
  \prod_{(i_1,i_2)\in\mathcal{E}}
  \frac{1}{\sqrt{\tau_{i_1i_2}^2}}
  \exp\left\{
  -\frac{\sum_{j=1}^p(a_{i_1j}-a_{i_2j})^2}{2\sigma^2\tau_{i_1i_2}^2}
  \right\}
  \\
  &&\quad\quad\times
  \prod_{(i_1,i_2)\in\mathcal{E}}
  \frac{\lambda^2_1w_{i_1i_2}^2}{2}
  \exp\left\{
  -\frac{\lambda^2_1w_{i_1i_2}^2}{2}\tau^2_{i_1i_2}
  \right\}
  \\
  &&\quad\quad\times
  \prod_{j=1}^p
  \frac{1}{\sqrt{\widetilde{\tau}_j^2}}
  \exp\left\{
  -\frac{\sum_{i=1}^n a_{ij}^2}{2\sigma^2\widetilde{\tau}_j^2}
  \right\}
  \\
  &&\quad\quad\times
  \prod_{j=1}^p
  \frac{\lambda^2_2 u_j^2}{2}
  \exp\left\{
  -\frac{\lambda^2_2 u_j^2}{2}\widetilde{\tau}_j^2
  \right\}
  \\
  &&\quad\quad\times
  \pi(\sigma^2)
  .
\end{eqnarray*}

The full conditional distribution is obtained as follows:
\begin{eqnarray*}
  \vector{a}_{\cdot j}|
  \vector{x}_{\cdot j},\{\tau_{i_1i_2}^2\}, \{\widetilde{\tau}_j^2\},\sigma^2
    &\sim&
      \mbox{N}_n(S^{-1}\vector{x}_{\cdot j},\ \sigma^2S^{-1}),
      \\
      &&
        S=S_\tau+(\widetilde{\tau}_j^{-2}+1)I_n,
  \\
  \frac{1}{\tau_{i_1i_2}^2}|
  \vector{a}_{i_1},\vector{a}_{i_2},\sigma^2
    &\sim&
      \mbox{IGauss}(\mu',\ \lambda'),
      \\
      &&
        \mu'=\frac{\sqrt{w_{i_1i_2}^2\lambda^2_1\sigma^2}}{
        \|\vector{a}_{i_1}-\vector{a}_{i_2}\|_2},
        \quad
        \lambda'=w_{i_1i_2}^2\lambda^2_1,
  \\
  \frac{1}{\widetilde{\tau}_j^2}|
  \vector{a}_{\cdot j},\sigma^2
    &\sim&
    \mbox{IGauss}(\widetilde{\mu}',\ \widetilde{\lambda}'),
    \\
    &&
      \widetilde{\mu}'=\frac{\sqrt{u_j^2\lambda_2^2\sigma^2}}{
      \|\vector{a}_{\cdot j}\|_2},
      \quad
      \widetilde{\lambda}'=u_j^2\lambda^2_2,
  \\
  \sigma^2|
  X,A,\{\tau_{i_1i_2}^2\}, \{\widetilde{\tau}_j^2\}
    &\sim&
      \mbox{IG}(\nu',\ \eta'),
      \\
      &&
        \nu'=np+\#\mathcal{E}+p+\nu_0,\\
        &&
        \eta'=
        \sum_{i=1}^n(\vector{x}_i-\vector{a}_i)^T(\vector{x}_i-\vector{a}_i)
        \\
        &&\qquad
        +\sum_{j=1}^p\vector{a}_{\cdot j}^T(S_\tau+\widetilde{\tau}_j^{-2}I_n)\vector{a}_{\cdot j}+
        \eta_0
  ,
\end{eqnarray*}
where $\mbox{IGauss}(x|\mu, \lambda)$ denotes the inverse-Gaussian distribution with density function
\begin{eqnarray*}
  \sqrt{\frac{\lambda}{2\pi}}
  x^{-3/2}
  \exp\left\{
    -\frac{\lambda(x-\mu)^2}{2\mu^2 x}
  \right\}
  ,\quad
  (x>0)
\end{eqnarray*}
and
\begin{eqnarray*}
S_\tau
&=&
\left\{
\begin{array}{cccc}
\sum_{1<i}\tau_{1i}^{-2}&-\tau_{12}^{-2}&\cdots&-\tau_{1n}^{-2}\\
-\tau_{12}^{-2}&\sum_{i<2}\tau_{i2}^{-2}+\sum_{2<i}\tau_{2i}^{-2}&\cdots&-\tau_{2n}^{-2}\\
\vdots&\vdots&\ddots&\vdots\\
-\tau_{1n}^{-2}&-\tau_{2n}^{-2}&\cdots&\sum_{i<n}\tau_{in}^{-2}
\end{array}
\right\}
.
\end{eqnarray*}

\subsection{Bayesian NEG sparse convex clustering\label{negscvc_gibbs}}

The prior distribution is transformed as follows:
\begin{eqnarray*}
  &&
  \pi(A,\{\tau_{i_1i_2}\},\{\psi_{i_1i_2}\},\{\widetilde{\tau}_j\},\{\widetilde{\psi}_j\},\sigma^2|X)
  \\
  &&\quad\propto
  (2\pi\sigma^2)^{-\frac{pn}{2}}
  \exp\left\{
  -\frac{1}{2\sigma^2}
  \sum_{i=1}^n
  (\vector{x}_i-\vector{a}_i)^T(\vector{x}_i-\vector{a}_i)
  \right\}
  \\
  &&\quad\quad\times
  (2\pi\sigma^2)^{-\#\mathcal{E}/2}
  \prod_{(i_1,i_2)\in\mathcal{E}}
  (\tau_{i_1i_2}^2)^{-1/2}
  \exp\left\{
     -\frac{\sum_{j=1}^p (a_{i_1j}-a_{i_2j})^2}{2\sigma^2\tau_{i_1i_2}^2}
  \right\}
  \\
  &&\quad\quad\times
  \prod_{(i_1,i_2)\in\mathcal{E}}
  \psi_{i_1i_2}\exp\{-\psi_{i_1i_2}\tau_{i_1i_2}^2\}
  \prod_{(i_1,i_2)\in\mathcal{E}}
  \frac{(\tau_{i_1i_2}^2)^{\lambda_1}}{\Gamma(\lambda_1)}
  \psi_{i_1i_2}^{\lambda_1-1}\exp\{-\gamma_1^2\psi_{i_1i_2}\}
  \\
  &&\quad\quad\times
  (2\pi\sigma^2)^{-p/2}
  \prod_{j=1}^p
  (\widetilde{\tau}_j^2)^{-1/2}
  \exp\left\{
    -\frac{\sum_{i=1}^n a_{ij}^2}{2\sigma^2\widetilde{\tau}_j^2}
  \right\}
  \\
  &&\quad\quad\times
  \prod_{j=1}^p
  \widetilde{\psi}_j\exp\{-\widetilde{\psi}_j\widetilde{\tau}_j^2\}
  \prod_{j=1}^p
  \frac{(\widetilde{\tau}_j^2)^{\lambda_2}}{\Gamma(\lambda_2)}
  \widetilde{\psi}_j^{\lambda_2-1}\exp\{-\gamma_2^2\widetilde{\psi}_j\}
  \\
  &&\quad\quad\times
  \pi(\sigma^2)
  .
\end{eqnarray*}

The full conditional distribution is obtained as follows:
\begin{eqnarray*}
  \vector{a}_{\cdot j}|
  \vector{x_{\cdot j}}, \{\tau_{i_1i_2}^2\}, \{\widetilde{\tau}_j^2\}, \sigma^2
    &\sim&
      \mbox{N}_n (S^{-1} \vector{x}_{\cdot j}, \sigma^2 S^{-1}),
      \\
      &&
        S=S_\tau+(\widetilde{\tau}_j^{-2}+1)I_n,
  \\
  \frac{1}{\tau_{i_1i_2}^2}|
  \vector{a}_{i_1}, \vector{a}_{i_2}, \psi_{i_1i_2}, \sigma^2
    &\sim&
      \mbox{IGauss}(\mu_{\tau_{i_1i_2}^{-2}},\lambda_{\tau_{i_1i_2}^{-2}}),
      \\
      &&
        \mu_{\tau_{i_1i_2}^{-2}} =  \frac{\sqrt{2\sigma^2\psi_{i_1i_2}}}{\|\vector{a}_{i_1}-\vector{a}_{i_2}\|_2 },
        \quad
        \lambda_{\tau_{i_1i_2}^{-2}} = 2\psi_{i_1i_2},
  \\
  \psi_{i_1i_2}|
  \tau_{i_1i_2}^2
    &\sim&
      \mbox{Ga}(k_{\psi_{i_1i_2}},\lambda_{\psi_{i_1i_2}}),
      \\
      &&
        k_{\psi_{i_1i_2}} = \lambda_1+1,
        \quad
        \lambda_{\psi_{i_1i_2}} = \tau_{i_1i_2}^2+\gamma_1^2,
  \\
  \frac{1}{\widetilde{\tau}_j^2}|
  \vector{a}_{\cdot j}, \widetilde{\psi}_j, \sigma^2
    &\sim&
      \mbox{IGauss}(\mu_{\widetilde{\tau}_j^{-2}},\lambda_{\widetilde{\tau}_j^{-2}}),
      \\
      &&
        \mu_{\widetilde{\tau}_j^{-2}} =  \frac{\sqrt{2\sigma^2\widetilde{\psi}_j}}{\|\vector{a}_{\cdot j}\|_2},
        \quad
        \lambda_{\widetilde{\tau}_j^{-2}} = 2\widetilde{\psi}_j,
  \\
  \widetilde{\psi}_j|
  \widetilde{\tau}_j^2
    &\sim&
      \mbox{Ga}(k_{\widetilde{\psi}_j}, \lambda_{\widetilde{\psi}_j}),
      \\
      &&
        k_{\widetilde{\psi}_j} = \lambda_2+1,
        \quad
        \lambda_{\widetilde{\psi}_j} = \widetilde{\tau}_j^2+\gamma_2^2,
  \\
  \sigma^2|
  X, A, \{\tau_{i_1i_2}^2\}, \{\widetilde{\tau}_j^2\}
    &\sim&
      \mbox{IG}(\nu',\eta'),
      \\
      &&
        \nu' = np + \#\mathcal{E} + p + \nu_0,
        \\
        &&
        \eta' =
        \sum_{i=1}^n(\vector{x}_i-\vector{a}_i)^t
        (\vector{x}_i-\vector{a}_i)\\
        &&\quad\quad + \sum_{j=1}^p \vector{a}_{\cdot j}^T (S_\tau+\widetilde{\tau}_j^{-2}I_n) \vector{a}_{\cdot j}^T
        + \eta_0,
\end{eqnarray*}
where
\begin{eqnarray*}
S_\tau
&=&
\left\{
\begin{array}{cccc}
\sum_{1<i}\tau_{1i}^{-2}&-\tau_{12}^{-2}&\cdots&-\tau_{1n}^{-2}\\
-\tau_{12}^{-2}&\sum_{i<2}\tau_{i2}^{-2}+\sum_{2<i}\tau_{2i}^{-2}&\cdots&-\tau_{2n}^{-2}\\
\vdots&\vdots&\ddots&\vdots\\
-\tau_{1n}^{-2}&-\tau_{2n}^{-2}&\cdots&\sum_{i<n}\tau_{in}^{-2}
\end{array}
\right\}
.
\end{eqnarray*}

\subsection{Bayesian horseshoe sparse convex clustering \label{horscvc_gibbs}}

The prior distribution is transformed as follows:

\begin{eqnarray*}
  &&
  \pi(A,\{\tau_{i_1i_2}\},\{\psi_{i_1i_2}\},\{\widetilde{\tau}_j\},\{\widetilde{\psi}_j\},\sigma^2|X)
  \\
  &&\quad\propto
  (2\pi\sigma^2)^{-\frac{pn}{2}}
  \exp\left\{
  -\frac{1}{2\sigma^2}
  \sum_{i=1}^n
  (\vector{x}_i-\vector{a}_i)^T(\vector{x}_i-\vector{a}_i)
  \right\}
  \\
  &&\quad\quad\times
  (2\pi\sigma^2\nu^2_1)^{-\frac{\#\mathcal{E}}{2}}
  \prod_{(i_1,i_2)\in\mathcal{E}}(\tau_{i_1i_2}^2)^{-1/2}
  \exp\left\{
    -\frac{1}{2\tau_{i_1i_2}^2\nu^2_1\sigma^2}
    \|\vector{a}_{i_1}-\vector{a}_{i_2}\|_2^2
  \right\}
  \\
  &&\quad\quad\times
  \prod_{(i_1,i_2)\in\mathcal{E}}
  \psi_{i_1i_2}^{-1/2}(\tau_{i_1i_2}^2)^{-1/2-1}
  \exp\left\{
    -\frac{1}{\psi_{i_1i_2}\tau_{i_1i_2}^2}
  \right\}
  \prod_{(i_1,i_2)\in\mathcal{E}}
  \psi_{i_1i_2}^{-1/2-1}
  \exp\left\{
    -\frac{1}{\psi_{i_1i_2}}
  \right\}
  \\
  &&\quad\quad\times
  (2\pi\sigma^2)^{-\frac{p}{2}}
  \prod_{j=1}^p(\widetilde{\tau}_j^2)^{-1/2}
  \exp\left\{
    -\frac{1}{2\sigma^2\widetilde{\tau}_j^2}
    \|\vector{a}_{\cdot j}\|_2^2
  \right\}
  \\
  &&\quad\quad\times
  \prod_{j=1}^p
  \widetilde{\psi}_j\exp\{-\widetilde{\psi}_j\widetilde{\tau}_j^2\}
  \prod_{j=1}^p
  \frac{(\gamma_2^2)^{\lambda_2}}{\Gamma(\lambda_2)}
  \widetilde{\psi}_j^{\lambda_2-1}\exp\{-\gamma_2^2\widetilde{\psi}_j\}
  \\
  &&\quad\quad\times
  \pi(\sigma^2)
  .
\end{eqnarray*}

The full conditional distribution is obtained as follows:
\begin{eqnarray*}
  \vector{a}_{\cdot j}|
  \vector{x}_{\cdot j}, \{\tau_{i_1i_2}^2\}, \{\widetilde{\tau}_j^2\}, \sigma^2
    &\sim&
      \mbox{N}_n(
        S^{-1}\vector{x}_{\cdot j},
        \sigma^2S^{-1}
      ),\\
    &&
      S=\frac{1}{\nu_1^2}S_\tau+
      \left(\frac{1}{\widetilde{\tau}_j^2}+1\right)
      I_n,
  \\
  \tau_{i_1i_2}^2|
  \vector{a}_{i_1}, \vector{a}_{i_2}, \psi_{i_1i_2}, \sigma^2
    &\sim&
    \mbox{IG}(\alpha_{\tau_{i_1i_2}^2},\beta_{\tau_{i_1i_2}^2}),
    \\
    &&
    \alpha_{\tau_{i_1i_2}^2}=1,
    \quad
    \beta_{\tau_{i_1i_2}^2}=
      \frac{1}{2\nu_1^2\sigma^2}
      \|\vector{a}_{i_1}-\vector{a}_{i_2}\|_2^2
      +\frac{1}{\psi_{i_1i_2}},
  \\
  \psi_{i_1i_2}|
  \tau_{i_1i_2}^2
    &\sim&
      \mbox{IG}(\alpha_{\psi_{i_1i_2}},\beta_{\psi_{i_1i_2}}),
      \\
      &&
      \alpha_{\psi_{i_1i_2}}=1,
      \quad
      \beta_{\psi_{i_1i_2}}=\frac{1}{\tau_{i_1i_2}^2}+1,
  \\
  \frac{1}{\widetilde{\tau}_j^2}|
  \vector{a}_{\cdot j}, \widetilde{\psi}_j, \sigma^2
    &\sim&
      \mbox{IGauss}(\mu_{\widetilde{\tau}_j^{-2}},\lambda_{\widetilde{\tau}_j^{-2}}),
      \\
      &&
      \mu_{\widetilde{\tau}_j^{-2}} =  \frac{\sqrt{2\sigma^2\widetilde{\psi}_j}}{\|\vector{a}_{\cdot j}\|_2},
      \quad
      \lambda_{\widetilde{\tau}_j^{-2}} = 2\widetilde{\psi}_j,
  \\
  \widetilde{\psi}_j|
  \widetilde{\tau}_j^2
    &\sim&
      \mbox{Ga}(k_{\widetilde{\psi}_j}, \lambda_{\widetilde{\psi}_j}),
      \\
      &&
      k_{\widetilde{\psi}_j} = \lambda_2+1,
      \quad
      \lambda_{\widetilde{\psi}_j} = \widetilde{\tau}_j^2+\gamma_2^2,
  \\
  \sigma^2|
  X, A, \{\tau_{i_1i_2}^2\}, \{\widetilde{\tau}_j^2\}
    &\sim&
      \mbox{IG}(\nu',\eta'),
      \\
      &&
      \nu' = np+\#\mathcal{E}+p+\nu_0,
      \\
      &&
      \eta' =
      \sum_{i=1}^n(\vector{x}_i-\vector{a}_i)^t
      (\vector{x}_i-\vector{a}_i)\\
      &&\quad\quad + \sum_{j=1}^p \vector{a}_{\cdot j}^T \left(\frac{1}{\nu_1^2}S_\tau+
      \frac{1}{\widetilde{\tau}_j^2}I_n\right) \vector{a}_{\cdot j}
      + \eta_0
  ,
\end{eqnarray*}
where
\begin{eqnarray*}
S_\tau
&=&
\left\{
\begin{array}{cccc}
\sum_{1<i}\tau_{1i}^{-2}&-\tau_{12}^{-2}&\cdots&-\tau_{1n}^{-2}\\
-\tau_{12}^{-2}&\sum_{i<2}\tau_{i2}^{-2}+\sum_{2<i}\tau_{2i}^{-2}&\cdots&-\tau_{2n}^{-2}\\
\vdots&\vdots&\ddots&\vdots\\
-\tau_{1n}^{-2}&-\tau_{2n}^{-2}&\cdots&\sum_{i<n}\tau_{in}^{-2}
\end{array}
\right\}
.
\end{eqnarray*}

\subsection{Bayesian Dirichlet--Laplace sparse convex clustering\label{dlscvc_gibbs}}

The prior distribution is transformed as follows:

\begin{eqnarray*}
  &&
  \pi(A,
  \{\tau_{i_1i_2}^2\},\{\psi_{i_1i_2}^2\},\nu,
  \{\widetilde{\tau}_{i_1i_2}^2\},\{\widetilde{\psi}_{i_1i_2}^2\},\widetilde{\nu},
  \sigma^2|X)
  \\
  &&\quad\propto
    (2\pi\sigma^2)^{-\frac{pn}{2}}
    \exp\left\{
    -\frac{1}{2\sigma^2}
    \sum_{i=1}^n
    (\vector{x}_i-\vector{a}_i)^T(\vector{x}_i-\vector{a}_i)
    \right\}
  \\
  &&\quad\quad\times
    \prod_{(i_1,i_2)\in\mathcal{E}}
    (2\pi\sigma^2\tau_{i_1i_2}^2\psi_{i_1i_2}\nu^2)^{-1/2}
    \exp\left\{
      -\frac{1}{2\sigma^2\tau_{i_1i_2}^2\psi_{i_1i_2}\nu^2}
      \|\vector{a}_{i_1}-\vector{a}_{i_2}\|_2^2
    \right\}
  \\
  &&\quad\quad\times
    \prod_{(i_1,i_2)\in\mathcal{E}}
    \frac{1}{2}\exp\left\{
      -\frac{1}{2}\psi_{i_1i_2}
    \right\}
    \prod_{(i_1,i_2)\in\mathcal{E}}
    \tau_{i_1i_2}^{\alpha_1-1}
  \\
  &&\quad\quad\times
    \nu^{\alpha_1\#\mathcal{E}-1}
    \exp\left\{-\frac{\nu}{2}\right\}
  \\
  &&\quad\quad\times
    (2\pi\sigma^2)^{-p/2}
    \prod_{j=1}^p
    (\widetilde{\tau}_j^2)^{-1/2}
    \exp\left\{
      -\frac{1}{2\sigma^2\widetilde{\tau}_j^2}
      \|\vector{a}_{\cdot j}\|_2^2
    \right\}
  \\
  &&\quad\quad\times
    \prod_{j=1}^p
    \widetilde{\psi}_j\exp\{-\widetilde{\psi}_j\widetilde{\tau}_j^2\}
    \prod_{j=1}^p
    \frac{(\gamma_2^2)^{\lambda_2}}{\Gamma(\lambda_2)}
    \widetilde{\psi}_j^{\lambda_2-1}\exp\{-\gamma_2^2\widetilde{\psi}_j\}
  \\
  &&\quad\quad\times
  \pi(\sigma^2)
  .
\end{eqnarray*}

The full conditional distribution is obtained as follows:
\begin{eqnarray*}
  \vector{a}_{\cdot j}|
  \vector{x}_{\cdot j}, \{\tau_{i_1i_2}^2\}, \{\psi_{i_1i_2}\}, \nu, \{\widetilde{\tau}_j^2\}, \sigma^2
    &\sim&
      \mbox{N}_n(
        S^{-1}\vector{x}_{\cdot j},
        \sigma^2S^{-1}
      ),
      \\
      &&
        S=
        \frac{1}{\nu^2}S_{\tau\psi}+
        \left(\frac{1}{\widetilde{\tau}_j^2}+1\right)
        I_n,
      \\
  T_{i_1 i_2}|
  \vector{a}_{i_1}, \vector{a}_{i_2}, \sigma^2
    &\sim&
      \mbox{giG}\left(
        \chi_{T_{i_1 i_2}},
        \rho_{T_{i_1 i_2}},
        \lambda_{T_{i_1 i_2}}
      \right),
      \\
      &&
        \chi_{T_{i_1 i_2}}=
        \frac{2\|\vector{a}_{i_1}-\vector{a}_{i_2}\|_2}{\sqrt{\sigma^2}}, \quad
        \rho_{T_{i_1 i_2}}=1, \quad
        \lambda_{T_{i_1 i_2}}=\alpha-1,
    \\
  \tau_{i_1i_2}
  &=&
    T_{i_1i_2} / \sum_{(i_1,i_2)\in\mathcal{E}} T_{i_1i_2},
    \\
  \frac{1}{\psi_{i_1i_2}}|
  \vector{a}_{i_1}, \vector{a}_{i_2}, \tau_{i_1i_2}, \nu, \sigma^2
    &\sim&
      \mbox{IGauss}(\mu_{\psi_{i_1i_2}^{-1}},\lambda_{\psi_{i_1i_2}^{-1}}),
      \\
      &&
        \mu_{\psi_{i_1i_2}^{-1}} =
        \frac{\nu\tau_{i_1i_2}\sqrt{\sigma^2}}{
        \|\vector{a}_{i_1}-\vector{a}_{i_2}\|_2}, \quad
        \lambda_{\psi_{i_1i_2}^{-1}} = 1,
    \\
  \nu|
  \vector{a}_{i_1}, \vector{a}_{i_2}, \tau_{i_1i_2}, \sigma^2
  &\sim&
  \mbox{giG}\left(
    \chi_{\nu},
    \rho_{\nu},
    \lambda_{\nu}
  \right),
  \\
  &&
  \chi_{\nu}=2\sum_{(i_1,i_2)\in\mathcal{E}}
  \frac{\|\vector{a}_{i_1}-\vector{a}_{i_2}\|_2}{\tau_{i_1i_2}\sqrt{\sigma^2}}, \quad
  \rho_{\nu}=1, \quad
  \lambda_{\nu}=(\alpha-1)\#\mathcal{E},
  \\
  \frac{1}{\widetilde{\tau}_j^2}|
  \vector{a}_{\cdot j}, \widetilde{\psi}_j, \sigma^2
    &\sim&
      \mbox{IGauss}(\mu_{\widetilde{\tau}_j^{-2}},\lambda_{\widetilde{\tau}_j^{-2}}),
      \\
      &&
      \mu_{\widetilde{\tau}_j^{-2}} =  \frac{\sqrt{2\sigma^2\widetilde{\psi}_j}}{\|\vector{a}_{\cdot j}\|_2},
      \quad
      \lambda_{\widetilde{\tau}_j^{-2}} = 2\widetilde{\psi}_j,
  \\
  \widetilde{\psi}_j|
  \widetilde{\tau}_j^2
    &\sim&
      \mbox{Ga}(k_{\widetilde{\psi}_j}, \lambda_{\widetilde{\psi}_j}),
      \\
      &&
      k_{\widetilde{\psi}_j} = \lambda_2+1,
      \quad
      \lambda_{\widetilde{\psi}_j} = \widetilde{\tau}_j^2+\gamma_2^2,
  \\
  \sigma^2|
  X, A, \{\tau_{i_1i_2}^2\}, \{\psi_{i_1i_2}\}, \nu, \{\widetilde{\tau}_j^2\}
    &\sim&
    \mbox{IG}(\nu',\eta'),
    \\
    &&
    \nu' = np+\#\mathcal{E}+p+\nu_0,
    \\
    &&
    \eta' =
    \sum_{i=1}^n(\vector{x}_i-\vector{a}_i)^t
    (\vector{x}_i-\vector{a}_i)\\
    &&\quad\quad + \sum_{j=1}^p \vector{a}_{\cdot j}^T \left(
    \frac{1}{\nu^2}S_{\tau\psi}+
    \frac{1}{\widetilde{\tau}_j^2}I_n
    \right) \vector{a}_{\cdot j}^T
    + \eta_0,
\end{eqnarray*}
where $\mbox{giG}\left(x|\chi,\rho,\lambda\right)$ is generalized inverse Gaussian
\begin{eqnarray*}
  z
  &\sim&
  \mbox{giG}(\chi,\rho,\lambda_0),
  \\
  \pi(z)
  &\propto&
  z^{\lambda_0-1}
  \exp\{
    -(\rho z+\chi/z)/2
  \},
\end{eqnarray*}
and
\begin{eqnarray*}
  S_{\tau\psi}
  =
  \left\{
  \begin{array}{cccc}
  \sum_{1<i_2}\tau_{1i_2}^{-2}\psi_{1i_2}^{-1} &
  -\tau_{12}^{-2}\psi_{12}^{-1} &
  \cdots &
  -\tau_{1n}^{-2}\psi_{1n}^{-1} \\
  -\tau_{12}^{-2}\psi_{12}^{-1} &
  \sum_{i_1<2}\tau_{i_12}^{-2}\psi_{i_12}^{-1}+\sum_{2<i_2}\tau_{2i_2}^{-2}\psi_{2i_2}^{-1} &
  \cdots & -\tau_{2n}^{-2}\psi_{2n}^{-1} \\
  \vdots & \vdots & \ddots & \vdots \\
  -\tau_{1n}^{-2}\psi_{1n}^{-1} &
  -\tau_{2n}^{-2}\psi_{2n}^{-1} &
  \cdots &
  \sum_{i_1<n}\tau_{i_1n}^{-2}\psi_{i_1n}^{-1}
  \end{array}
  \right\}
  .
\end{eqnarray*}

\begin{acknowledgements}
S. K. was supported by JSPS KAKENHI Grant Number JP19K11854 and MEXT KAKENHI Grant Numbers JP16H06429, JP16K21723, and JP16H06430.
Super-computing resources were provided by Human Genome Center (the Univ. of Tokyo).
\end{acknowledgements}

%
%

\bibliographystyle{apalike}
\bibliography{main}

\end{document}